\documentclass[review]{elsarticle}
\usepackage[misc]{ifsym}

\usepackage{times}
\usepackage{latexsym}
\usepackage{microtype}
\usepackage{xcolor}
\usepackage{graphicx}
\usepackage{booktabs}
\usepackage{enumerate}
\usepackage{amsmath}
\usepackage{amsfonts}
\usepackage{mathtools}
\usepackage{amssymb}
\usepackage{amsthm}
\usepackage{relsize}
\usepackage[linesnumbered,ruled,vlined]{algorithm2e}
\usepackage{algpseudocode}
\usepackage[inline]{enumitem}
\usepackage{soul}
\usepackage{multirow}
\usepackage{caption}
\usepackage{subcaption}
\setlist[itemize]{noitemsep, topsep=0pt}
\usepackage[nodisplayskipstretch]{setspace}

\usepackage{lineno}

\definecolor{darkgreen}{rgb}{0.0, 0.2, 0.13}

\newcommand{\eg}{\emph{e.g.}}
\newcommand{\ie}{\emph{i.e.}}
\newcommand{\etal}{\emph{et al.}}
\newcommand{\topic}[1]{\hfill\break\noindent\textbf{#1}.}

\newcommand{\ourmodel}{\AE MP}
\newcommand{\weights}{\textrm{\textbf{W}}}
\newcommand{\loss}{\textrm{\textbf{L}}}
\newcommand{\kgpath}[3]{$#1\xrightarrow{#2}#3$}

\journal{Knowledge-based Systems}

\begin{document}

\begin{frontmatter}



\title{Learning Attention-based Representations from Multiple Patterns for Relation Prediction in Knowledge Graphs}


\author{V\'{i}tor Louren\c{c}o\corref{vitor}}
\ead{vitorlourenco@id.uff.br}
\cortext[vitor]{Corresponding author}

\author{Aline Paes}
\ead{alinepaes@ic.uff.br}

\address{Institute of Computing, Universidade Federal Fluminense\\
Av Gal Milton Tavares de Souza, S/N, Boa Viagem, \\
Niter\'{o}i, RJ, Brazil}

\begin{abstract}
  Knowledge bases, and their representations in the form of knowledge graphs~(KGs), are naturally incomplete. Since scientific and industrial applications have extensively adopted them, there is a high demand for solutions that complete their information. Several recent works tackle this challenge by learning embeddings for entities and relations, then employing them to predict new relations among the entities. Despite their aggrandizement, most of those methods focus only on the local neighbors of a relation to learn the embeddings. As a result, they may fail to capture the KGs' context information by neglecting long-term dependencies and the propagation of entities' semantics. In this manuscript, we propose \ourmodel{}~(\textbf{A}ttention-based \textbf{E}mbeddings from \textbf{M}ultiple \textbf{P}atterns), a novel model for learning contextualized representations by: \emph{(i)} acquiring entities context information through an attention-enhanced message-passing scheme, which captures the entities local semantics while focusing on different aspects of their neighborhood; and \emph{(ii)} capturing the semantic context, by leveraging the paths and their relationships between entities. Our empirical findings draw insights into how attention mechanisms can improve entities' context representation and how combining entities and semantic path contexts improves the general representation of entities and the relation predictions. Experimental results on several large and small knowledge graph benchmarks show that \ourmodel{} either outperforms or competes with state-of-the-art relation prediction methods.
\end{abstract}




\end{frontmatter}


\section{Introduction\label{sec:1}}
Humankind's knowledge processed by machines can be posed as sets of information (\textit{i.e.}, facts, beliefs, and rules) circa entities. By connecting these sets of information and their relationships, we can build a network of knowledge. Knowledge bases~(KBs) play a fundamental role in compiling this network of knowledge to structure and store information, properties, and relationships.

Existing compiled KBs~\cite{miller1995wordnet,bollacker2008freebase} have succeeded as core resources on language and knowledge-related tasks such as question answering~\cite{yih2015qa}, information retrieval~\cite{liu2018entity}, recommender systems~\cite{palumbo2017entity2rec}, and even commonsense reasoning~\cite{lin2019kagnet}, among others. Typically, the information in these bases are organized in the form of triples \mbox{$(head, relation, tail)$}, in which $head$ and $tail$ are entities (also called \textit{subject} and \textit{object}, respectively) and $relation$ (also called \textit{predicate}) is the relationship between both entities that qualifies the triple semantics. KBs are commonly represented as \emph{knowledge graphs} (KG), which are labeled multi-digraph whose nodes are the entities ($head$ and $tail$), the edges express relationships between entities, and the edges' labels define the semantics of each relationship ($relation$).

Despite the success of KGs as core resources for a large plethora of tasks, real-world KGs usually experience incompleteness and noisy information. They are strictly dependable on new information acquisition, which is often hard to obtain and noisy prone once the information continuously evolves over time~\cite{levesque1984ikb}. Thus, an essential need regarding KGs arises: the existence of automatic methods to complete their information. This capacity involves solving the prediction of missing relations among entities in the KG. Such a task is either framed as \textit{i)}~link or entity prediction, when the target is to find a missing entity, given the relation and the other entity (\mbox{$(head, relation, ?)$}, for example); or \textit{ii)}~relation prediction when the target is to find a missing relation (\mbox{$(head, ?, tail)$}). Often, KGs suffer from completing missing facts given existing ones (relation prediction), differently from question-answering-related tasks, which already have a pair of entity-relationship aiming to find the other correspondent entity (entity prediction)~\cite{cui2021tarp,wang2021harp}. Consequently, targeting completing missing facts in KGs, in this manuscript, we follow the formalization of the \emph{relation prediction} task {\footnote{Although in this manuscript we make such a distinction, there are previous works that use those two terms interchangeably. For example, in~\cite{nathani-etal-2019-learning} and and~\cite{zhang2022ar}, the proposed methods focus on what we call link prediction but nominate the task as relation prediction.}}. 

Several previous works have proposed techniques to predict and infer new relationships targeting completing knowledge graphs. Such methods can be roughly divided into two categories~\cite{dumancic2019srl}: \textit{i)}~distributional~\cite{bordes2013transe,trouillon2016complex,sun2018rotate,zhang2019quate,xu-li-2019-relation} and \textit{ii)}~symbolic methods~\cite{galarraga2015amie,wang2016knowledge}. The former relies on encoding the interactions between entities' neighborhoods into low-dimensional dense vectors, \ie{}, embeddings vectors, which allow for new relationships between entities to be predicted from interactions of the embeddings~\cite{nickel2016kge}. Differently, the later~\cite{galarraga2015amie,wang2016knowledge} aim at predicting new relations from the observed examples in the KG by extracting logical patterns in the form of semantic paths between entities~\cite{nickel2016kge}. Methods grounded in the symbolic paradigm can reason over complex relational paths due to their rule-based grounding. However, they are well-known for suffering from scalability issues, which limit their rule-inference search space. In contrast, the embedding-based methods scale, but they can only learn local structures, requiring further enhancements for more generality. Section~\ref{sec:2} provides more details on previous work focused on completing KGs.

\begin{figure}[t]
    \centering
    \includegraphics[width=0.7\linewidth]{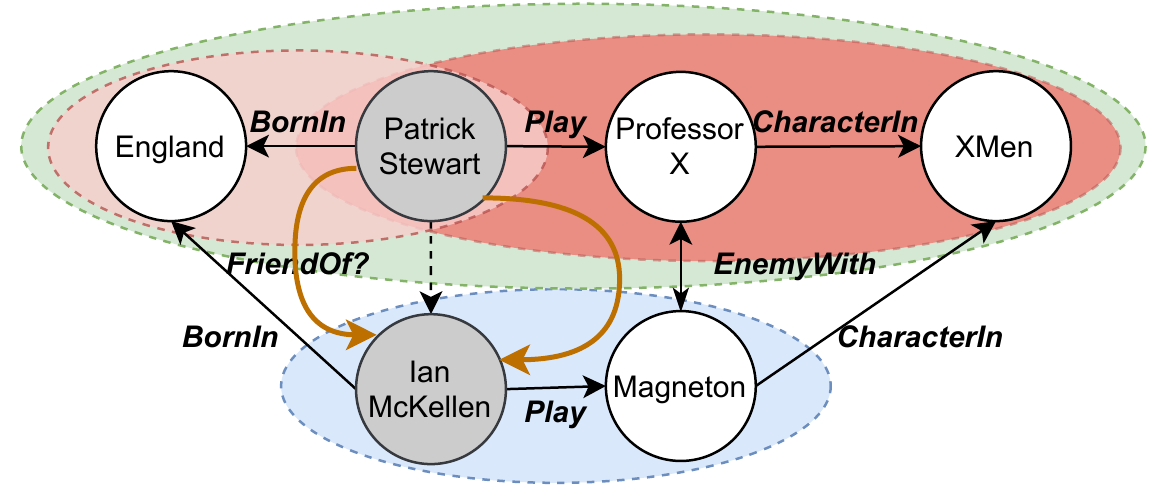}
    \caption{Example of a knowledge graph. Gray circles indicate the head and tail entities. The dashed arrow is a potentially missing relation. Orange arrows indicate the semantic paths between the head and tail, each one taking a distinct path (\textsl{BornIn-FriendOf and Play-EnemyWith-Play}). Green, red, and blue-shaded areas represent the global, local, and random contexts, respectively, that may be considered by each attention mechanism of~\ourmodel{}.}
    \label{fig:example}
\end{figure}

In this manuscript, we address the relation prediction challenge to complete knowledge bases designing an embedding-based model called \ourmodel{}~(\textbf{A}ttention-based \textbf{E}m\-beddings from \textbf{M}ultiple \textbf{P}atterns). \ourmodel{} aims at the major advantages of each paradigm: efficient embeddings learning from the distributional paradigm and capturing generality with semantic paths from the symbolic paradigm. To that, \ourmodel{} addresses learning contextualized representations from the combination of entities' context and semantic paths. The key insights of \ourmodel{} are based on the perception that multiple patterns exist to relate entities. Like a word in a sentence, entities' local neighborhood evidences their structure and properties, while semantic paths relate to long-term dependencies. This contextual information's unification enhances information to predict new relations among existing entities in knowledge graphs. 

Figure~\ref{fig:example} exemplifies the patterns and mechanisms that \ourmodel{} uses to learn representations. Primarily, \ourmodel{} operates an attention-enhanced message-passing scheme to iteratively propagate the $k$-hop local neighbor information of a given entity over the neighborhood edges, paying attention in local (Figure~\ref{fig:example} red-shaded areas -- light red is the 1-hop neighbors, and the dark red is the 2-hop neighbors), global (Figure~\ref{fig:example} green area), and random (Figure~\ref{fig:example} blue area) aspects of the neighborhood to learn a unique contextualized representation of the given entity and its neighborhood. Secondly, \ourmodel{} identifies the paths between a pair of entities and combines them into a single semantic path representation (Figure~\ref{fig:example} yellow arrows). Finally, \ourmodel{} combines both contextualized entities and semantic path representations in order to inform the probability of a new relationship between a pair of entities. Section~\ref{sec:3} presents \ourmodel{} in great detail.

We conduct an extensive evaluation of \ourmodel{} on the main KGs benchmarks (WN18~\cite{bordes2013transe}, WN18RR~\cite{dettmers2017wn18rr}, FB15k~\cite{bordes2013transe}, and FB15k-237~\cite{toutanova2015fb15k237}). Furthermore, we experiment with two small KGs~\cite{kok2007umlskinship} (UMLS and Kinship) to assess \ourmodel{} when only limited information is available. We compare \ourmodel{}'s efficiency against state-of-the-art models in the relation prediction task. The results show that \ourmodel{} outperforms the compared methods in five (WN18, WN18RR, FB15k, UMLS, and Kinship) out of the six datasets and demonstrates competitive results on the sixth benchmark. Also, we conduct several ablation studies, where we assess the importance of each pattern observed by the model, namely, the semantic paths and their length, the number of entities context hops on overall performance, and the learning capabilities enabled by each attention mechanism and their combinations. All the results are presented and described in Section~\ref{sec:4}.

The main contributions of this manuscript are:
\begin{enumerate}[leftmargin=*]
    \setlength{\parskip}{0pt}
    \setlength{\parsep}{0pt}
    \item[i)] An attention-enhanced message-passing scheme for learning contextualized entities representations;
    \item[ii)] Identification of semantic paths between a pair of entities and the combination of the paths' representations into a single general semantic path representation; 
    \item[iii)] Combination of contextualized entities representations and semantic path representations for relation prediction; and
    \item[iv)] Extensive experimental evaluation on several knowledge graphs benchmarks showing the proposed solutions either outperforms or competes with existing state-of-the-art methods.
\end{enumerate}

\begin{figure}[t]
    \centering
    \includegraphics[width=\linewidth]{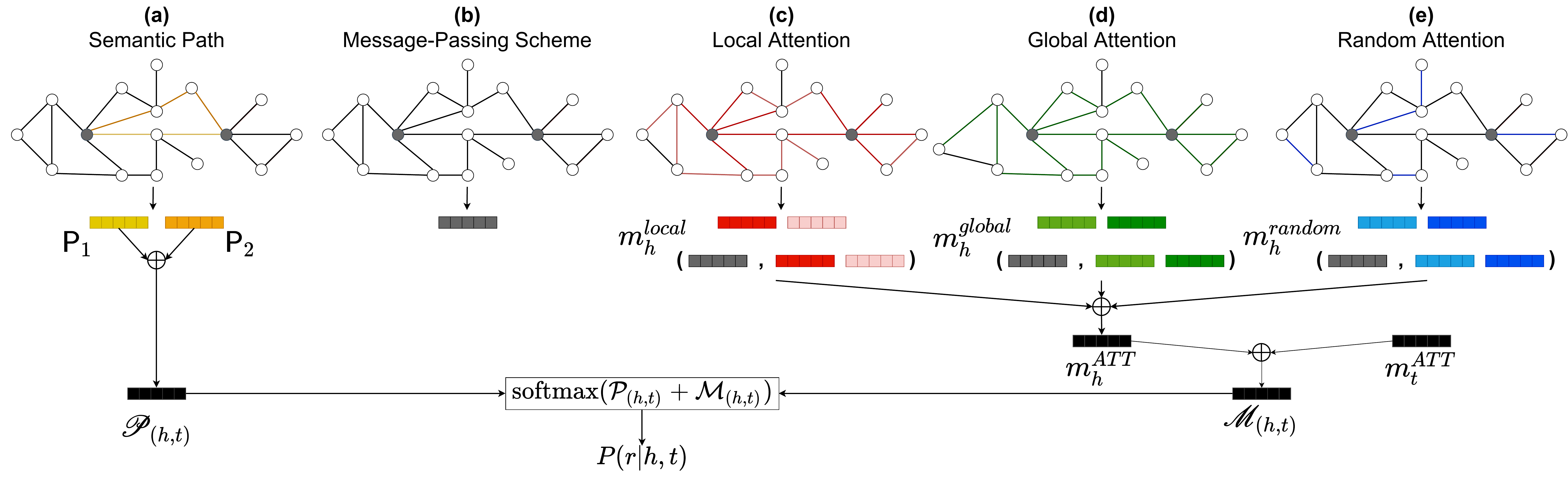}
    \caption{Overview of \ourmodel{} architecture. Boxes represent the embeddings produced in each step. Yellow-shaded elements are related to the semantic paths. Gray-shaded illustrates the message-passing scheme. Red-shaded points out each hop considered within the local attention mechanisms. Green-shaded expresses the iterations associated with the global attention mechanism. Blue-shaded indicates the randomly selected relationships used within random attention. Light and dark shades of the same color illustrate different neighborhood representations of entity $h$, which are captured and combined by the color-associated mechanism forming the entity $m_{\texttt{\scriptsize{h}}}^{\texttt{\scriptsize{mechanism}}}$.}
    
    \label{fig:model}
\end{figure}

\section{Related work\label{sec:2}}

Most current distributional methods for learning KGs induce entity embeddings based only on their direct neighbors~\cite{bordes2013transe,trouillon2016complex,dettmers2017wn18rr}, neglecting that the triples where entities appear are part of a much richer structure (the KG itself). In the following, we revisit methods with similar motivation as ours: entities are likely to be better represented when contextual information is incorporated into the learning process; hence, one may reach better predictive results to complete the KGs. We also review some recent approaches that rely on attention mechanisms over knowledge graphs representation. Finally, as \ourmodel{} incorporates the message-passing mechanism, commonly used in graph neural networks, we also bring attention to some of those methods in this section.

\topic{Entities Context}
Luo~\etal{}~\cite{luo2015lccp} pioneered the generation of context-dependent entity embeddings by learning contextual connectivity patterns from 1-hop neighborhood of entities. Later, Oh~\etal{}~\cite{oh2018calc} proposed context-aware embeddings by jointly learning from an entity and its multi-hop neighborhood. Both approaches only adopt the entities' neighbors' representations as context, neglecting the semantics of the relationships between neighbors. In contrast, we build \ourmodel{} upon Wang~\etal{}~\cite{wang2020pathcon}, which addressed entities' local context from a message-passing scheme to learn entities $k$-hop neighborhood based on the aggregation of graph's edges. However, although the message-passing scheme holds the potential to acquire entities' local structures, \cite{wang2020pathcon} equally weights the neighbors within the aggregation process. Thus, in this paper, we also benefit from the message-passing scheme, yet we adopt an attention mechanism to focus on different aspects of the passed messages. With that, we weight the importance of each fact in the neighborhood for the final entity representation.

\topic{Relational Paths}
Several of previous works count with semantic composition among entities to add contextual information to KG's embeddings, either in the form of constraints or as sequence of relations~\cite{lin2015ptranse,guu2015traversing,das2016chains,ding2018improving,guo2019learning,lin2019relation}. PTransE~\cite{lin2015ptranse} extends the use of translation-based embeddings mechanisms~\cite{bordes2013transe} by including multiple-step relation paths between two entities. Gu~\etal{}~\cite{guu2015traversing} proposes additive and multiplicative compositions over relations while~\cite{neelakantan2015compositional} leverages recurrent neural networks to consider relational paths of entities. Aiming at complex reasoning to populate KBs from texts, ~\cite{toutanova2015representing} and \cite{das2016chains} incorporate multiple paths built not only upon relations but also with other entities. More recently, Guo~\etal{}~\cite{guo2019learning} also targeted at learning from relational paths by using a recurrent neural network with residual connections. 
\ourmodel{} also builds paths motivated by the semantic enrichment that one entity may provide to the others. However, we focus on message-passing schemas leveraged by different attention patterns to let the model find out during the learning which elements are more relevant to the embeddings and, consequently, the relation prediction task.


\topic{Attention-based Embeddings}
GAT~\cite{velickovic2018gat} employs the concept of attention to learn representations from graph-based data. It computes attention weights over every other node (in the most general formulation) while not considering possible labeled relations between them. Here, besides allowing different views of attention over the nodes and edges, we aim at predicting relations in \emph{knowledge} graphs. 
The first attempt to use attention-based mechanisms to predict missing entities in KGs was presented in \cite{nathani-etal-2019-learning}. To capture distant entities' contributions, they relied on a relation composition mechanism that introduces auxiliary edges between neighbors in hops. Also, attention mechanisms are used to encode additional information within the learned representation~\cite{wang2021harp,zhang2022ar}. \ourmodel{}, on the other hand, allows for local and global attention mechanisms to potentially focus on close and distant neighbors during the message-passing mechanism. Moreover, we include a random attention mechanism that potentially learns \emph{when} to attend near and distant neighbors to predict missing relations. Selecting neighbors at random to proceed with the message passing introduces an indiscriminate bias once it bases itself on the sparse aspects of the node's neighborhood.

\topic{Graph Neural Networks}
Commonly, Graph Neural Networks (GNN) methods follow the propagation-aggregation node-based message-passing paradigm over multi-hop neighbors~\cite{scarselli2008graph,kipf2017semi,hamilton2017inductive,xu2019dynamically}. \ourmodel{} also follows the message-passing schema, yet the messages are passed along the edges to capture semantic relationships expressed on them. The entities' messages are aggregated from the relation representations that connect them. Previous approaches have also explored GNN to complete knowledge graphs, primarily focusing on link/entity prediction tasks. Several of them leverage on the local neighborhood context of entities only~\cite{schlichtkrull2018modeling,bansal2019a2n,nathani-etal-2019-learning,zhang2020relational} while here we explore global and random contexts besides the local neighborhood. There are many ways to explore local information. For instance,~\citet{zhang2020relational} assume there is a hierarchical structure in the local neighborhood of an entity while ~\citet{schlichtkrull2018modeling} and ~\cite{shang2019end} extend Convolutional Neural Network to knowledge base completion by addressing the local context of nodes. ~\citet{teru2020inductive} also take advantage of a GNN framework yet assuming an entity-independent semantics to learn to predict relations to handle inductive reasoning problems. They built the learning framework upon several other approaches and offered a set of inductive relation prediction benchmarks to the community. Evolving \ourmodel{} to the inductive setting, particularly regarding the semantic paths component, is an appealing direction for the future.

\section{Our Approach: Attention-based Embeddings from Multiple Patterns\label{sec:3}}
In this section we propose \ourmodel{}\footnote{https://github.com/MeLLL-UFF/AEMP}~(\textbf{A}ttention-based \textbf{E}mbeddings from \textbf{M}ultiple \textbf{P}atterns). Figure~\ref{fig:model} depicts an illustration of how the different components of the proposed architecture interact. Roughly speaking, \ourmodel{} includes entities' context enhanced by three attention mechanisms and semantic paths to induce the embeddings of entities. Those embeddings are the input of a softmax function that defines the probability of a relation connects two entities, \ie{}, a (missing) candidate triple. We start by defining the problem we tackle. Next, we provide \ourmodel{} components in details.

\begin{table}[h]
\centering
\caption{Notations used in this paper.}
\label{tab:notation}
\begin{tabular}{@{}ll@{}}
\toprule
Symbol & Description \\ \midrule
\multicolumn{1}{l|}{$h$, $t$} & Head and tail entities\\
\multicolumn{1}{l|}{$r$} & Relationship between a pair of entities \\
\multicolumn{1}{l|}{$m_{e}^{i}$} & Message of entity $e$ at iteration $i$ \\ 
\multicolumn{1}{l|}{$s_{r}^{i}$} & Representation of a relationship $r$ at iteration $i$\\ 
\multicolumn{1}{l|}{$\mathcal{N}(e)$} & Incident relationships of an entity $e$ \\ 
\multicolumn{1}{l|}{$\mathcal{N}(r)$} & Incident entities to a relationship $r$ \\ 
\multicolumn{1}{l|}{$\weights$} & Arbitrary weights matrix \\ 
\multicolumn{1}{l|}{$\lambda$} & Attention mechanism (local, global, or random) \\ 
\multicolumn{1}{l|}{$\alpha_{rk}^{\lambda}$} & Attention alignment score \\ 
\multicolumn{1}{l|}{${s^{\lambda}}_{r}^{k}$} & Attention-enhanced relationship representation \\ 
\multicolumn{1}{l|}{$\mathcal{CR}_{r}$} & Context set of the relationship $r$ \\ 
\multicolumn{1}{l|}{$\mathcal{C}_{I}$} & Iterations' context set \\ 
\multicolumn{1}{l|}{$m_{e}^{ATT}$} & Final representation of an entity $e$ \\ 
\multicolumn{1}{l|}{$\mathcal{M}_{(h,t)}$} & Final entities context representation of entities pair $(h,t)$ \\ 
\multicolumn{1}{l|}{$\mathcal{P}_{(h,t)}$} & Final semantic path representation of entities pair $(h,t)$ \\ 
\bottomrule
\end{tabular}
\end{table}

\topic{Problem Formulation}\label{sec:problem}
Given a Knowledge Base represented by a labeled multi-digraph $\mathcal{KG} = (\mathcal{E}, \mathcal{R}, \mathcal{F})$, where $\mathcal{E}$ is the set of nodes that represents the entities, $\mathcal{R}$ is the set of edges' labels that represents the relations, and $\mathcal{F}: \mathcal{E} \times \mathcal{R} \times \mathcal{E}$ is the set of triples $(h, r, t)$ that represent logical facts $r(h,t)$. Our goal is, given a pair of entities $(h,t)$, where $h \in \mathcal{E}$ and $t \in \mathcal{E}$, to predict the existence of a relation $r$ connecting these two entities given by the probability $P(r|h,t)$. The outcome assembles a new (missing) triple $f' = (h, r, t)$, where $r \in \mathcal{R}$ is the predicted relation, and $f'$ is the candidate triple. The notations used in this section and the rest of this manuscript, together with their descriptions, are presented in Table~\ref{tab:notation}.


\subsection{Entities Context}\label{sec:entcont}
Here we describe the attention-enhanced mechanism for learning the joint representation of entities and their contexts, according to a message-passing scheme (overview in Figure~\ref{fig:model}.(b)). \ourmodel{} employs three attention mechanisms to learn the entities's representation, namely, a local (Figure~\ref{fig:model}.(c)), global (Figure~\ref{fig:model}.(d)), and random (Figure~\ref{fig:model}.(e)) attention.

Algorithm~\ref{alg:entcon} brings the steps proposed here for capturing and learning entities' context. The development of this attention-enhanced procedure is inspired by the advances in attention mechanisms over graphs (GATs)~\cite{velickovic2018gat}, as well as their sparse versions~\cite{martins2016softmax,zaheer2020bigbird}. The message-passing scheme resembles the aggregating entities' neighborhood representations proposed in~\cite{wang2020pathcon}. \ourmodel{} relies on the message-passing scheme to interactively learn the entities' representations from the propagation of their multi-hop neighborhood edges representations. After, it submits the learned representations to an attention layer that combines local, global, and random attention mechanisms. Such a combination of different views benefits the model to learn the entities' context while focusing on different aspects of the neighborhood. Following the example illustrated in Figure~\ref{fig:example}, local attention reinforces local connective patterns, providing a narrow view of the entities and their surroundings (\eg{}, the red-shaded areas in the figure; the relationship between entities \texttt{PatrickStewart} and \texttt{England}). Global attention reinforces global connective patterns, broadly focusing on the entities neighborhood (\eg{}, the green-shaded area in the figure; the relationship between entities \texttt{England} and \texttt{XMen}). Random attention assists in capturing non-directed patterns (\eg{}, the blue-shaded area in the figure; the relationship between entities \texttt{IanMcKellen} and \texttt{Magneton}).

\setlength{\textfloatsep}{0pt}
\begin{algorithm}[ht]
\small
\DontPrintSemicolon
\SetAlgoLined
\caption{Attention-enhanced Message-Passing Scheme}\label{alg:entcon}
\textbf{Input: maximum number of iterations $K$, maximum hop $H$, probability for the random mechanism $p$} \\
\textbf{Output: final representation message $M^{ATT}$} \\
$k \gets 1$\; 
$S_{i} \gets S_{0}$\; 
$S^{random} \gets \{\}$\; 
$S^{global} \gets S_{0}$\; 
$S^{local} \gets S_{0}$\;
\While{$k \neq K$}{
    $S_{k+1} \gets \{\}$\;
    $hop \gets 1$\;
    \While{$hop \neq H$}{
        $M_{k} \gets \textrm{Equation}\ref{eq:mpsm}(S_{i})$\; 
  
        $s_{r}^{k+1} \gets\textrm{Equation}\ref{eq:mpss}(M_{k})$\;
        $S_{k+1} \gets S_{k+1} \cup s_{r}^{k+1}$\;
        $S^{local} \gets S^{local} \cup s_{r}^{k+1}$\;
        \uIf{with a probability $p$}{
            $S^{random} \gets S^{random} \cup s_{r}^{k+1}$\;
        }
        $hop \gets hop + 1$\;
    }
    $S^{local} \gets \textrm{Equation}\ref{eq:localatt}(S_{k+1}, S^{local})$\;
    $S^{global} \gets S^{global} \cup S_{k+1}$\;
    $k \gets k + 1$\;
}

${M^{local}} \gets S^{local}$\; 

$M^{global} \gets \textrm{Equation}\ref{eq:globalatt}(S_{K}, S^{global})$\; 
$M^{random} \gets \textrm{Equation}\ref{eq:randomatt}(S_{K}, S^{random})$\; 

$M^{ATT} \gets M^{local} \oplus M^{global} \oplus M^{random}$\; 

\Return $M^{ATT}$\;
\end{algorithm}

Equations~\ref{eq:mpsm} calculates the message representation of an entity $e$ according to the massage-passing scheme, while Figure~\ref{fig:model}.(b) illustrates it. The message $m_{e}^{i} \in \mathbb{R}^D$ is the contextual representation of an entity $e \in \mathcal{E}$, acording to the context captured as $s_{r}^{i}$, which is the representation of a relationship $r$ (an edge) between a pair of entities:
\begin{equation}\label{eq:mpsm}\small
    m_{e}^{i} = \sum_{r \in \mathcal{N}(e)} s_{r}^{i}, 
\end{equation}

Both $m_{e}^{i}$ and $s_{r}^{i}$ are the entity message and relationship representation computed in the $i$-th iteration of the outer loop of the Algorithm~\ref{alg:entcon}. Note that the message associated with an entity $e$ is built upon the aggregation of the relationship's representation that connects $e$ to another entity, according to $e$'s neighborhood $\mathcal{N}(e)$. The messages of two entities are combined through a pairwise product: 


\setlength{\abovedisplayskip}{0pt} \setlength{\abovedisplayshortskip}{0pt}

\setlength{\abovedisplayskip}{0pt} \setlength{\abovedisplayshortskip}{0pt}
\begin{equation}\label{eq:mpss}\small
    \begin{split}
    s_{r}^{i+1} = \sigma\left(\textrm{flatten}\left({m_{h}^{i}m_{t}^{i}}^\intercal\right)\;\weights_{1}^{i} + s_{r}^{i}\;\weights_{2}^{i} + b^{i}\right),\\~\\
    h,t \in \mathcal{N}(r),\; m_{h}^{i}m_{t}^{i} = \begin{bmatrix}
        m_{h}^{i(1)}m_{t}^{i(1)} & \dots & m_{h}^{i(1)}m_{t}^{i(d)} \\
        & \ddots & \\
        m_{h}^{i(d)}m_{t}^{i(1)} & \dots & m_{h}^{i(d)}m_{t}^{i(d)}
    \end{bmatrix},
    \end{split}
\end{equation}
\setlength{\belowdisplayskip}{0pt} \setlength{\belowdisplayshortskip}{0pt}

Precisely, the entity's message representation $m_{e}^{i} \in \mathbb{R}^D$ calculated within an iteration $i$ is the sum of all current incident relationship representation $s_r^{i}$ where $r \in \mathcal{N}(e)$ (the neighbourhood of the entity $e$) (Equation~\ref{eq:mpsm}). The relation representation is also updated iteratively according to Equation~\ref{eq:mpss}, where the next relation representation $s_{r}^{i+1}$ is updated based on a cross-neighbor aggregator operation~\cite{wang2020pathcon}. The aggregator models the entities' cross matrix, a pairwise product of the entities' message representations $m_{h}^{i(d_{v})}m_{t}^{i(d_{v})}$, where $m_{h}^{i}$ and $m_{t}^{i}$ are the head and tail representations, respectively, $r$ connects $h$ and $t$, and $(d_v)$ is each dimension in $\mathbb{R}^D$. The updated relation representation $s_{r}^{i+1}$ is a combination of the flattened entities' cross matrix with the current relation representation $s_{r}^{i}$, both linear transformed with learned weights $\weights_1$ and $\weights_2$, followed by a final nonlinear transformation $\sigma$.

Built on top of the message-passing scheme, we define a multi-context attention layer motivated by GAT and the sparse attention mechanism~\cite{zaheer2020bigbird}. We extend the GATs' attention mechanism by adopting local and global attention. Moreover, we include random attention within the message-passing scheme iterations to leverage from sparsely arranged possible relationships.

Assume a particular iteration $1 \leq k \leq K$, where $K$ is the maximum number of iterations according to Algorithm~\ref{alg:entcon}. The local attention mechanism focuses on the relationships' messages acquired in \emph{each hop} visited within $k$ (inner loop of Algorithm~\ref{alg:entcon}) as the context message $s^{k}$ resultant of that iteration. The context message allows for the querying message to be informed of its local context structure captured by its neighborhood structural information. Figure~\ref{fig:model}.(c) illustrates this mechanism and Equations~\ref{eq:localatta} and~\ref{eq:localatt} formalize it:  

\setlength{\belowdisplayskip}{0pt} \setlength{\belowdisplayshortskip}{0pt}
\begin{equation}\label{eq:localatta}\small
    \alpha_{rk}^{local} = \textrm{align}_{r}(s^{k}, s_{r}^{k}) = \frac{\textrm{exp}({s^{k}}^\intercal \;\weights_{l_\alpha}\; s_{r}^{k})}{\sum_{r' \in \mathcal{C}_{R}}\textrm{exp}({s^{k}}^\intercal \;\weights_{l_\alpha}\; s_{r'}^{k})},
\end{equation}
\setlength{\abovedisplayskip}{0pt} \setlength{\abovedisplayshortskip}{0pt}
\begin{equation}\label{eq:localatt}\small
    {s^{local}}_{r}^{k} = \sigma\left(\sum_{r' \in \mathcal{C}_{R}}\alpha_{r'k}^{local} \;\weights_{s_\alpha}\; s_{r'}^{k}\right),
\end{equation}
\setlength{\belowdisplayskip}{0pt} \setlength{\belowdisplayshortskip}{0pt}

\noindent where $\alpha_{rk}^{local}$ is the local attention weight calculated at iteration $k$ when focusing on a relation $r$, $s^{k}$ is the gathered representation of relations participating at the iteration $k$, $s_{r}^{k}$ is the representation of a particular relation $r$ at the iteration $k$, $\mathcal{CR}_{r}$ is the set of other relations in the relation $r$'s local context, \ie{}, those relations sharing a node with $r$, and $\{\weights_{l_\alpha}, \weights_{l_s}\} \in \mathbb{R}^{D~\times~D}$ are arbitrary weights associated with the local attention. Finally, ${s^{local}}_{r}^{k}$ is the local relation representation. To calculate it, \ourmodel{} first calculates a linear composition of the local attention weights and representation of relations that belong to the context $\mathcal{CR}_{r}$ of $r$, weighted by a learned matrix $\weights_{s_\alpha}$. After, to finally get ${s^{local}}_{r}^{k}$, that linear transformation passes through a nonlinear transformation with $\sigma$.

The global attention, illustrated in (Figure~\ref{fig:model}.(d), operates over the final entities' messages of each iteration. By doing so, the global semantic of each iteration is captured and informed to the querying message. The following equations define how \ourmodel{} computes the global attention weights and related relation representations:

\setlength{\belowdisplayskip}{0pt} \setlength{\belowdisplayshortskip}{0pt}
\begin{equation}\label{eq:globalatta}\small
    \alpha_{rk}^{global} = \textrm{align}_{k}(s_{r}, s_{r}^{k}) = \frac{\textrm{exp}({s_{r}}^\intercal \;\weights_{g_\alpha}\; s_{r}^{k})}{\sum_{i \in \mathcal{C}_{I}}\textrm{exp}({s_{r}}^\intercal \;\weights_{g_\alpha}\; s_{r}^{i})},
\end{equation}
\setlength{\belowdisplayskip}{0pt} \setlength{\belowdisplayshortskip}{0pt}
\setlength{\abovedisplayskip}{0pt} \setlength{\abovedisplayshortskip}{0pt}
\begin{equation}\label{eq:globalatt}\small
    {s^{global}}_{r}^{k} = \sigma\left(\sum_{i \in \mathcal{C}_{I}}\alpha_{ri}^{global} \weights_{g_s}\; s_{r}^{i}\right),
\end{equation}
\setlength{\belowdisplayskip}{0pt} \setlength{\belowdisplayshortskip}{0pt}

\noindent where $\alpha_{rk}^{global}$ are the global attention weights, $s_r$ is the representation of the relation $r$ detached from any iteration, while $s_{r}^{k}$ is the representation of the relation $r$ computed within the iteration $k$, $\{\weights_{g_\alpha} \weights_{g_s}\} \in \mathbb{R}^{D~\times~D}$, $\weights_g$ are learned according to the global attention, and $\mathcal{C}_{I}$ is the set of relations taken into account during the iteration $k$.

Note that the local attention has the point of view of  representations computed at a iteration $k$, namely $s^k$ and $s_r^k$ while the global attention also focuses on $s_r^k$ but aligns it with the representation of the relation detached from a specific iteration, namely, $s_r$.

Finally, the random attention, illustrated in Figure~\ref{fig:model}.(e), randomly captures further aspects of the entities' neighborhood. Those are information that might get neglected by the local attention based on each hop or the global attention based on a whole iteration update. In this way, the random attention weights are built upon both the hops and iterations. It acts only if a probability $p$ is reached, as depicted in lines $13$ and $14$ of Algorithm~\ref{alg:entcon}, precisely to capture different information than the local and global mechanisms. In this way, the random mechanism allows the querying message to pay attention to sparse aspects of the node's neighborhood, introducing an indiscriminate bias. While the relations representations are updated at each iteration in the outermost loop for the global and local mechanisms, the random context is only updated if a randomly generated number reaches the probability $p$.

The random attention mechanism is formalized as follows: 

\setlength{\belowdisplayskip}{0pt} \setlength{\belowdisplayshortskip}{0pt}
\begin{equation}\label{eq:randomatta}\small
    \alpha_{rk}^{random} = \textrm{align}_{rk}(s, s_{r}^{k}) = \frac{\textrm{exp}(s^\intercal \;\weights_{r_\alpha}\; s_{r}^{k})}{\sum_{\{i,r'\} \in \mathcal{CR}_{r}}\textrm{exp}(s^\intercal \;\weights_{r_\alpha}\; s_{r'}^{i})},
\end{equation}
\setlength{\belowdisplayskip}{0pt} \setlength{\belowdisplayshortskip}{0pt}
\setlength{\abovedisplayskip}{0pt} \setlength{\abovedisplayshortskip}{0pt}
\begin{equation}\label{eq:randomatt}\small
    {s^{random}}_{r}^{k} = \sigma\left(\sum_{\{i,r\} \in \mathcal{CR}_{r}}\alpha_{ri}^{random} \;\weights_{r_s}\; s_{r}^{i}\right),
\end{equation}
\setlength{\belowdisplayskip}{0pt} \setlength{\belowdisplayshortskip}{0pt}

\noindent where $\alpha_{rk}^{random}$ are the random attention weights, $s$ is a representation captured from any relation, $s_{r}^{k}$ is the representation of the relation $r$ at the iteration $k$,  $\mathcal{CR}_{r}$ is the context of a relation, $\{\weights_{r_\alpha} \weights_{r_s}\} \in \mathbb{R}^{D~\times~D}$, $\weights_r$ are learned to weigh the random attention weights.

Each attention mechanism provides a final representation from their point of view to an entity $e$. In this way, the message representation $m_{e}^{local}$ of an entity $e$ according to the local attention is given by

\setlength{\abovedisplayskip}{0pt} \setlength{\abovedisplayshortskip}{0pt}
\begin{equation}\label{eq:melocal}\small
    m_{e}^{local} = \sum_{r \in \mathcal{N}(e)} s^{local}_{r}, 
\end{equation}
\setlength{\belowdisplayskip}{0pt} \setlength{\belowdisplayshortskip}{0pt}

\noindent by summing out the local representations of each relation $r$ in the neighbourhood $\mathcal{N}(e)$ of $e$, where the local representations are computed according to Equation~\ref{eq:localatt} after reaching the final iteration when $k=K$.

Next, the message representation $m_{e}^{global}$ of an entity $e$ according to the global attention attention mechanism is given by
\setlength{\abovedisplayskip}{0pt} \setlength{\abovedisplayshortskip}{0pt}
\begin{equation}\label{eq:meglobal}\small
    m_{e}^{global} = \sum_{r \in \mathcal{N}(e)} {s^{global}}_{r}, 
\end{equation}
\setlength{\belowdisplayskip}{0pt} \setlength{\belowdisplayshortskip}{0pt}\textbf{}

\noindent computed over similar conditions of the local message.

Finally, the message find out by the random attention mechanism is defined by the following equation, with similar definitions as above. 

\setlength{\abovedisplayskip}{0pt} \setlength{\abovedisplayshortskip}{0pt}
\begin{equation}\label{eq:merandom}\small
    m_{e}^{random} = \sum_{r \in \mathcal{N}(e)} {s^{random}}_{r}. 
\end{equation}
\setlength{\belowdisplayskip}{0pt} \setlength{\belowdisplayshortskip}{0pt}

The final representation of an entity and its local, global, and random contexts ${m_{e}^{ATT}}$ is the concatenation of the outcome representation of each attention mechanism:

\setlength{\abovedisplayskip}{0pt} \setlength{\abovedisplayshortskip}{0pt}
\begin{equation}\small
    m_{e}^{ATT} = m_{e}^{local} \oplus m_{e}^{global} \oplus m_{e}^{random},
\end{equation}
\setlength{\belowdisplayskip}{0pt} \setlength{\belowdisplayshortskip}{0pt}

Finally, to provide the final embedding of the head and tail pair $\mathcal{M}_{(h,t)}$, we concatenate the head and tail final messages:

\setlength{\abovedisplayskip}{0pt} \setlength{\abovedisplayshortskip}{0pt}
\begin{equation}\small
    \mathcal{M}_{(h,t)} = m_{h}^{ATT} \oplus m_{t}^{ATT},
\end{equation}
\setlength{\belowdisplayskip}{0pt} \setlength{\belowdisplayshortskip}{0pt}

where $m_{h}^{ATT}$ and $m_{t}^{ATT}$ are the final representation messages from entities head and tail, respectively.

\subsection{Semantic Path Context}\label{sec:pathcont}
Next, we describe the capture of semantic paths (Figure~\ref{fig:model}.(a)) and how they are aimed as context information within the learning process. 
The aforementioned attention-based approaches provide essential information for learning entities' local context. However, it may fail at capturing long semantic paths, \ie{}, long sequences of relations between entities. As an example, the semantic path (Figure~\ref{fig:example} yellow-shaded arrows) between \texttt{PatrickStewart} and \texttt{IanMcKellen} entities is overlooked, since it is longer than the entities context hops. To address this issue, \ourmodel{} introduces representations learned from  the semantic paths between entities when learning the embeddings to predict a missing relation.

\begin{figure}[t]
    \centering
    \includegraphics[width=0.7\linewidth]{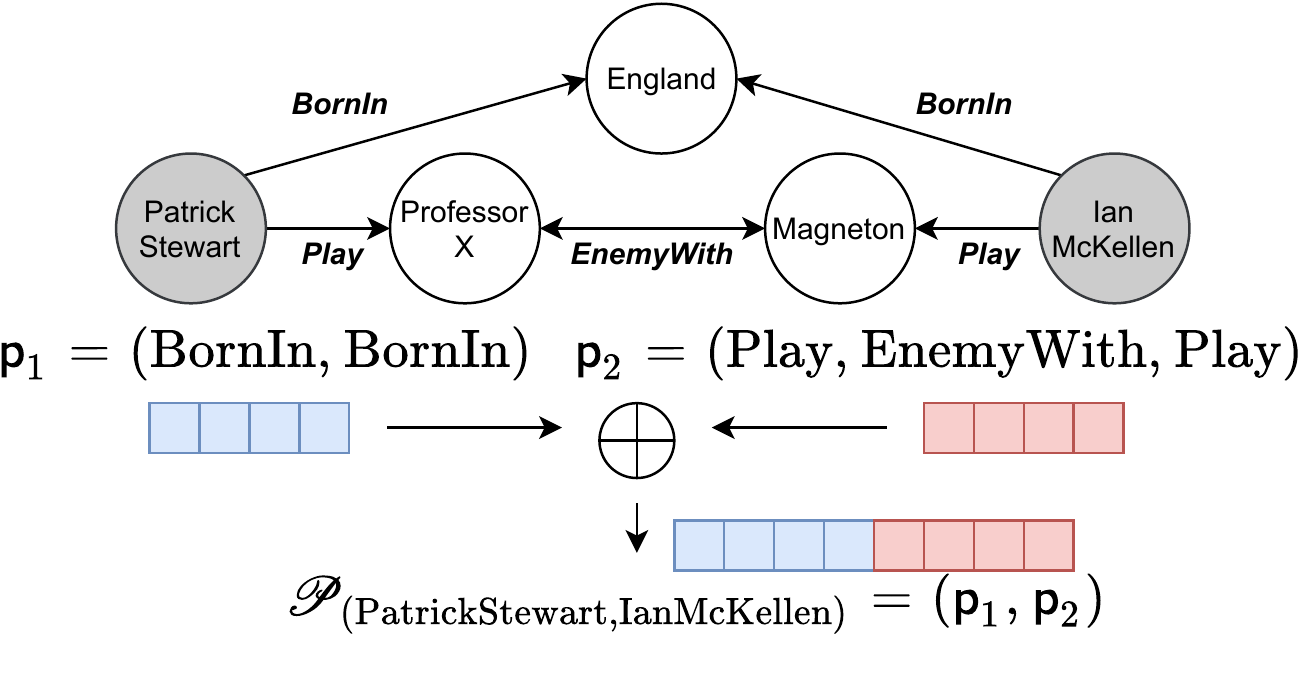}
    \caption{Example of the semantic paths' representation between head and tail entities, \ie{}, \texttt{PatrickStewart} and \texttt{IanMcKellen} entities, respectively.}
    \label{fig:path}
\end{figure}

Algorithm~\ref{alg:algo2} describes the procedure for capturing the semantic paths and  Figure~\ref{fig:path} illustrates the mechanism. The procedure first identifies $m$ semantic paths $\mathsf{P}(h,t) = \{\mathsf{p}_{1}, \dots, \mathsf{p}_{m}\}$ between two entities (Algorithm~\ref{alg:algo2} -- Line 1), where $h$ and $t$ are the head and tail entities.  $\mathsf{p}_{i} = (r_{1}, \dots, r_{in})$ is a semantic path  \kgpath{h}{r_{1},\dots,r_{in}}{t} of size $in$ between the entities $h$ and $t$, and $r_{j}$ is a relation within the path. The procedure collect the paths with a depth-limited breadth-first search algorithm constrained to a maximum path length. After, the algorithm creates one-hot representations to the paths (Algorithm~\ref{alg:algo2} -- Lines 2-4) and, to learn a single representation of the semantic paths, it performs a linear transformation~\cite{lin2015ptranse} (Algorithm~\ref{alg:algo2} -- Line 6) over the concatenation of the paths' representations (Algorithm~\ref{alg:algo2} -- Line 4), mapping the concatenated representation to the same dimensional space of $\mathcal{M}_{(h,t)}$:

\setlength{\abovedisplayskip}{0pt} \setlength{\abovedisplayshortskip}{0pt}
\begin{equation}\label{eq:paths}\small
    \mathcal{P}_{(h,t)} = \weights_p\; \bigg\|_{\mathsf{p} \in \mathsf{P}(h,t)} \mathsf{P}(h,t),
\end{equation}
\setlength{\belowdisplayskip}{0pt} \setlength{\belowdisplayshortskip}{0pt}
\noindent where $\weights_p$ denotes the linear transformation matrix weights, and $\|$ denotes the concatenation operation.

\setlength{\textfloatsep}{0pt}
\begin{algorithm}[t]
\small
\DontPrintSemicolon
\SetAlgoLined
\caption{Learning semantic paths representation}\label{alg:algo2}
$\mathsf{P}(h,t) \gets BreadthFirstSearch(h,t,max\_length)$\;
$\mathcal{P}_{(h,t)} \gets OneHotEncoding(\mathsf{p}'); \; \mathsf{p}' \in \mathsf{P}(h,t)$

\ForEach {$\mathsf{p} \in \mathsf{P}(h,t) - \{\mathsf{p}'\}$}{
    $\mathcal{P}_{(h,t)} \gets \mathcal{P}_{(h,t)} \oplus OneHotEncoding(\mathsf{p})$
}

$\mathcal{P}_{(h,t)} \gets \weights\;\mathcal{P}_{(h,t)} $

\Return $\mathcal{P}_{(h,t)}$\;
\end{algorithm}

\subsection{Training Objective}\label{sec:trainingobj}

To predict relations over KGs, we rely on a loss function based on the distribution probability of a relation over a head and tail pair~\cite{wang2020pathcon}. Like so, we take the probability distribution computed by a softmax function from the addition of both entities context $\mathcal{M}_{(h,t)}$ and semantic path context $\mathcal{P}_{(h,t)}$ representations:

\setlength{\belowdisplayskip}{0pt} \setlength{\belowdisplayshortskip}{0pt}
\begin{equation}\small
    P(r|h,t) = \text{softmax}(\mathcal{M}_{(h,t)} + \mathcal{P}_{(h,t)}).
\end{equation}

We target at minimizing the cross-entropy loss between the predicted probability of a training fact $\mathcal{T}$ and its ground-truth:

\setlength{\belowdisplayskip}{0pt} \setlength{\belowdisplayshortskip}{0pt}
\begin{equation}\label{eq:loss}\small
    \textrm{min}\;\loss(\Omega) = - \sum_{(h,r,t) \in \mathcal{T}} r\,\textrm{log}(P(r|h,t; \Omega)).
\end{equation}

\section{Experiments\label{sec:4}}
We conduct experiments on real-world KGs largely used in the literature and report the results compared with several previous methods that have also focused on the relation prediction task. Furthermore, we conduct three ablation studies, where we draw insights over \ourmodel{} and its variations.


\subsection{Experimental Setting\label{sec:expsetting}}

\topic{Datasets} The experiments are based on six widely used datasets: WN18, WN18RR, FB15k, FB15k-237, UMLS, and Kinship. Details and statistics about them can be found in~\ref{sec:appA}. WN18~\cite{bordes2013transe} is a subset from WordNet~\cite{miller1995wordnet}, a KG containing lexical relations among English words. FB15k~\cite{bordes2013transe} is a subset from Freebase~\cite{bollacker2008freebase}, a KG containing human knowledge facts. WN18RR~\cite{dettmers2017wn18rr} and FB15k-237~\cite{toutanova2015fb15k237} are subsets of WN18 and FB15k, respectively, where the original inverse relations were removed. UMLS~\cite{kok2007umlskinship} is built upon data from the Unified Medical Language System~\cite{mccray2003umls}, a biomedical ontology. Kinship~\cite{kok2007umlskinship} contains information regarding kinship relationships among members of an Australian tribe~\cite{denham1973kinship}. We used the originally provided data splits for WN18, WN18RR, FB15k, and FB15k-237, and for UMLS and Kinship, we use the data splits provided by~\cite{nathani-etal-2019-learning}.

\topic{Baselines} 
When experimenting with WN18, WN18RR, FB15k, and FB15k-237, we compare \ourmodel{}, and its variations, with six models that have results also targetting at the relation prediction task: TransE~\cite{bordes2013transe}, ComplEx~\cite{trouillon2016complex}, SimplE~\cite{kazemi2018simple}, RotatE~\cite{sun2018rotate}, QuatE~\cite{zhang2019quate} are the models representing the state-of-the-art in embedding-based models, while PathCon~\cite{wang2020pathcon} is the state-of-the-art closer to our model, using entities context and relational path as features in the learning process. We reused the results reported in~\cite{wang2020pathcon}. We trained from scratch UMLS, and Kinship with PathCon~\cite{wang2020pathcon} as this is the most similar method to \ourmodel{} and the SOTA in the relation prediction task.

\topic{Implementation details} We implemented \ourmodel{}\footnote{https://github.com/MeLLL-UFF/AEMP} using PyTorch~\cite{NEURIPS2019_9015} and trained using a single Nvidia V100 GPU. \ref{sec:appB} shows the hyperparameters' search space, which were varied in the experiments. 
In our best experimentation settings, we have a learning rate of $10^{-3}$ with Adam~\cite{kingma2015adam} optimizer. To avoid overfitting, we employ L2 regularization using L2 weight loss of $10^{-7}$. We adopt a batch size of 128, 25 training epochs, hidden states of 64 dimensions, and 0.2 as the random attention context selection criteria. Finally, on WN18, WN18RR, UMLS, and Kinship benchmarks, we employed three-hops entities context and semantic path length up to three; on FB15k and FB15k-237 benchmarks, we employed two-hops entities' context and semantic path length up to two, due to hardware limitations and the size of each benchmark.

\topic{Evaluation protocol} We evaluate all models on their capabilities of completing missing facts on KGs, assessing them under the \emph{relation prediction} task. As described in Section~\ref{sec:problem}, the task aims to infer a new fact $f = (h,r,t)$ by predicting a relation $r$ given a pair of entities $(h,t)$. We report results from Mean Reciprocal Rank (MRR), Mean Rank (MR), and correctly predicted relations (Hit ratio) in the top 1 and 3 rank evaluation metrics. A lower value of MR points out better results, while the other metrics target higher values. The results are the average and standard deviation values from five independent executions.

\begin{table*}[t]
\caption{Relation prediction results on WN18 and WN18RR datasets. $L$, $G$, and $R$ stand for the local, global, and random attention mechanisms, respectively. [*]: Results are taken from~\cite{wang2020pathcon}. The best and second-best results are highlighted in bold and underlined, respectively.}
\centering
\resizebox{\textwidth}{!}{%
\begin{tabular}{@{}c|cccc|cccc@{}}
\toprule
Model & \multicolumn{4}{c|}{WN18} & \multicolumn{4}{c}{WN18RR} \\
 & MRR $\uparrow$ & MR $\downarrow$& Hit@1 $\uparrow$& Hit@3 $\uparrow$& MRR $\uparrow$& MR $\downarrow$& Hit@1 $\uparrow$& Hit@3 $\uparrow$\\ \midrule
TransE* & 0.971 & 1.160 & 0.955 & 0.984 & 0.784 & 2.079 & 0.669 & 0.870 \\
ComplEx* & 0.985 & 1.098 & 0.979 & 0.991 & 0.840 & 2.053 & 0.777 & 0.880 \\
SimplE* & 0.972 & 1.256 & 0.964 & 0.976 & 0.730 & 3.259 & 0.659 & 0.755 \\
RotatE* & 0.984 & 1.139 & 0.979 & 0.986 & 0.799 & 2.284 & 0.735 & 0.823 \\
QuatE* & 0.981 & 1.170 & 0.975 & 0.983 & 0.823 & 2.404 & 0.767 & 0.852 \\
PathCon & 0.9915 \footnotesize{$\pm$ 0.0007} & \underline{ 1.0275 \footnotesize{$\pm$ 0.0039}} & 0.9859 \footnotesize{$\pm$ 0.0010} & \underline{ 0.9970 \footnotesize{$\pm$ 0.0007}} & 0.9689 \footnotesize{$\pm$ 0.0025} & 1.0839 \footnotesize{$\pm$ 0.0088} & 0.9447 \footnotesize{$\pm$ 0.0038} & 0.9935 \footnotesize{$\pm$ 0.0013} \\ \midrule
\ourmodel~(L) & 0.9450 \footnotesize{$\pm$ 0.0526} & 1.1503 \footnotesize{$\pm$ 0.1144} & 0.8991 \footnotesize{$\pm$ 0.0983} & 0.9958 \footnotesize{$\pm$ 0.0011} & 0.9632 \footnotesize{$\pm$ 0.0134} & 1.1257 \footnotesize{$\pm$ 0.0312} & 0.9366 \footnotesize{$\pm$ 0.0283} & 0.9876 \footnotesize{$\pm$ 0.0054} \\
\ourmodel~(G) & 0.9917 \footnotesize{$\pm$ 0.0006} & \textbf{1.0274 \footnotesize{$\pm$ 0.0018}} & 0.9863 \footnotesize{$\pm$ 0.0008} & 0.9969 \footnotesize{$\pm$ 0.0004} & 0.9764 \footnotesize{$\pm$ 0.0014} & \textbf{1.0645 \footnotesize{$\pm$ 0.0047}} & 0.9580 \footnotesize{$\pm$ 0.0024} & 0.9942 \footnotesize{$\pm$ 0.0009} \\
\ourmodel~(R) & 0.9908 \footnotesize{$\pm$ 0.0008} & 1.0282 \footnotesize{$\pm$ 0.0021} & 0.9845 \footnotesize{$\pm$ 0.0014} & \textbf{0.9971 \footnotesize{$\pm$ 0.0004}} & 0.9710 \footnotesize{$\pm$ 0.0002} & 1.0773 \footnotesize{$\pm$ 0.0034} & 0.9480 \footnotesize{$\pm$ 0.0008} & \underline{ 0.9948 \footnotesize{$\pm$ 0.0009}} \\
\ourmodel~(L+G) & \textbf{0.9942 \footnotesize{$\pm$ 0.0004}} & 1.0288 \footnotesize{$\pm$ 0.0037} & \textbf{0.9920 \footnotesize{$\pm$ 0.0005}} & 0.9952 \footnotesize{$\pm$ 0.0003} & \underline{ 0.9792 \footnotesize{$\pm$ 0.0019}} & 1.0717 \footnotesize{$\pm$ 0.0079} & \underline{ 0.9659 \footnotesize{$\pm$ 0.0027}} & 0.9916 \footnotesize{$\pm$ 0.0023} \\
\ourmodel~(L+R) & 0.9717 \footnotesize{$\pm$ 0.0101} & 1.1020 \footnotesize{$\pm$ 0.0343} & 0.9532 \footnotesize{$\pm$ 0.0164} & 0.9904 \footnotesize{$\pm$ 0.0056} & \textbf{0.9807 \footnotesize{$\pm$ 0.0035}} & 1.0736 \footnotesize{$\pm$ 0.0100} & \textbf{0.9677 \footnotesize{$\pm$ 0.0074}} & 0.9908 \footnotesize{$\pm$ 0.0021} \\
\ourmodel~(G+R) & 0.9907 \footnotesize{$\pm$ 0.0007} & 1.0297 \footnotesize{$\pm$ 0.0037} &  0.9846 \footnotesize{$\pm$ 0.0010} & 0.9965 \footnotesize{$\pm$ 0.0005} & 0.9758 \footnotesize{$\pm$ 0.0029} & \underline{ 1.0652 \footnotesize{$\pm$ 0.0074}} & 0.9568 \footnotesize{$\pm$ 0.0049} & \textbf{0.9949 \footnotesize{$\pm$ 0.0011}} \\
\ourmodel~(L+G+R) & \underline{ 0.9940 \footnotesize{$\pm$ 0.0005}} & 1.0336 \footnotesize{$\pm$ 0.0075} & \underline{ 0.9917 \footnotesize{$\pm$ 0.0007}} & 0.9949 \footnotesize{$\pm$ 0.0006} & 0.9786 \footnotesize{$\pm$ 0.0010} & 1.0679 \footnotesize{$\pm$ 0.0047} & 0.9635 \footnotesize{$\pm$ 0.0018} & 0.9937 \footnotesize{$\pm$ 0.0011} \\ \bottomrule
\end{tabular}%
}
\label{tab:results1}
\end{table*}

\begin{table*}[t]
\centering
\caption{Relation prediction results on FB15k and FB15k-237 datasets. $L$, $G$, and $R$ stand for the local, global, and random attention mechanisms, respectively. [*]: Results are taken from~\cite{wang2020pathcon}. The best and second-best results are highlighted in bold and underlined, respectively.}
\resizebox{\textwidth}{!}{%
\begin{tabular}{@{}c|cccc|cccc@{}}
\toprule
Model & \multicolumn{4}{c|}{FB15k} & \multicolumn{4}{c}{FB15k-237} \\
 & MRR $\uparrow$& MR $\downarrow$& Hit@1 $\uparrow$& Hit@3 $\uparrow$& MRR $\uparrow$& MR $\downarrow$& Hit@1 $\uparrow$& Hit@3 $\uparrow$\\ \midrule
TransE* & 0.962 & 1.684 & 0.940 & 0.982 & 0.966 & 1.352 & 0.946 & 0.984 \\
ComplEx* & 0.901 & 1.553 & 0.844 & 0.952 & 0.924 & 1.494 & 0.879 & 0.970 \\
SimplE* & 0.983 & 1.308 & 0.972 & 0.991 & 0.971 & 1.407 & 0.955 & 0.987 \\
RotatE* & 0.979 & 1.206 & 0.967 & 0.986 & 0.970 & 1.315 & 0.951 & 0.980 \\
QuatE* & \textbf{0.984} & 1.207 & 0.972 & 0.991 & 0.974 & 1.283 & 0.958 & 0.988 \\
PathCon & 0.9821 \footnotesize{$\pm$ 0.0002} & 1.5115 \footnotesize{$\pm$ 0.0585} & \underline{ 0.9699 \footnotesize{$\pm$ 0.0003}} & \textbf{0.9940 \footnotesize{$\pm$ 0.0002}} & \textbf{0.9797 \footnotesize{$\pm$ 0.0005}} & \textbf{1.1588 \footnotesize{$\pm$ 0.0214}} & \textbf{0.9653 \footnotesize{$\pm$ 0.0009}} & \textbf{0.9944 \footnotesize{$\pm$ 0.0004}} \\ \midrule
\ourmodel~(L) & 0.9689 \footnotesize{$\pm$ 0.0017} & 1.0823 \footnotesize{$\pm$ 0.0060} & 0.9445 \footnotesize{$\pm$ 0.0028} & 0.9876 \footnotesize{$\pm$ 0.0022} & 0.7900 \footnotesize{$\pm$ 0.0546} & 2.8834 \footnotesize{$\pm$ 0.4814} & 0.7188 \footnotesize{$\pm$ 0.0664} & 0.8296 \footnotesize{$\pm$ 0.0649} \\
\ourmodel~(G) & \underline{ 0.9824 \footnotesize{$\pm$ 0.0004}} & 1.4948 \footnotesize{$\pm$ 0.0637} & \textbf{0.9704 \footnotesize{$\pm$ 0.0005}} & \textbf{0.9940 \footnotesize{$\pm$ 0.0003}} & \underline{ 0.9790 \footnotesize{$\pm$ 0.0007}} & 1.1950 \footnotesize{$\pm$ 0.0326} & 0.9640 \footnotesize{$\pm$ 0.0012} & \underline{ 0.9943 \footnotesize{$\pm$ 0.0004}} \\
\ourmodel~(R) & 0.9815 \footnotesize{$\pm$ 0.0004} & 1.4466 \footnotesize{$\pm$ 0.0806} & 0.9688 \footnotesize{$\pm$ 0.0006} & \underline{ 0.9938 \footnotesize{$\pm$ 0.0002}} & \textbf{0.9797 \footnotesize{$\pm$ 0.0004}} & \underline{ 1.1922 \footnotesize{$\pm$ 0.0255}} & \underline{ 0.9652 \footnotesize{$\pm$ 0.0006}} & 0.9941 \footnotesize{$\pm$ 0.0004} \\
\ourmodel~(L+G) & 0.9798 \footnotesize{$\pm$ 0.0015} & \textbf{1.0666 \footnotesize{$\pm$ 0.0095}} & 0.9664 \footnotesize{$\pm$ 0.0021} & 0.9917 \footnotesize{$\pm$ 0.0029} & 0.9765 \footnotesize{$\pm$ 0.0011} & 1.3631 \footnotesize{$\pm$ 0.0941} & 0.9643 \footnotesize{$\pm$ 0.0017} & 0.9877 \footnotesize{$\pm$ 0.0003} \\
\ourmodel~(L+R) & 0.9032 \footnotesize{$\pm$ 0.0136} & 3.2039 \footnotesize{$\pm$ 0.2885} & 0.8714 \footnotesize{$\pm$ 0.0161} & 0.9202 \footnotesize{$\pm$ 0.0144} & 0.8724 \footnotesize{$\pm$ 0.0436} & 2.7244 \footnotesize{$\pm$ 0.5824} & 0.8369 \footnotesize{$\pm$ 0.0545} & 0.8885 \footnotesize{$\pm$ 0.0446} \\
\ourmodel~(G+R) & 0.9823 \footnotesize{$\pm$ 0.0002} & 1.4866 \footnotesize{$\pm$ 0.0478} & \textbf{0.9704 \footnotesize{$\pm$ 0.0004}} & \underline{ 0.9938 \footnotesize{$\pm$ 0.0003}} & \underline{ 0.9790 \footnotesize{$\pm$ 0.0004}} & 1.2103 \footnotesize{$\pm$ 0.0442} & 0.9643 \footnotesize{$\pm$ 0.0007} & 0.9940 \footnotesize{$\pm$ 0.0004} \\ 
\ourmodel~(L+G+R) & 0.9786 \footnotesize{$\pm$ 0.0020} & \underline{ 1.0695 \footnotesize{$\pm$ 0.0075}} & 0.9641 \footnotesize{$\pm$ 0.0039} & 0.9924 \footnotesize{$\pm$ 0.0027} & 0.9754 \footnotesize{$\pm$ 0.0011} & 1.4346 \footnotesize{$\pm$ 0.0467} & 0.9622 \footnotesize{$\pm$ 0.0018} & 0.9877 \footnotesize{$\pm$ 0.0006} \\ \bottomrule
\end{tabular}%
}
\label{tab:results2}
\end{table*}

\begin{table*}[t]
\centering
\caption{Relation prediction results on UMLS and Kinship datasets. $L$, $G$, and $R$ stand for the local, global, and random attention mechanisms, respectively. The best and second-best results are highlighted in bold and underlined, respectively.}
\label{tab:results3}
\resizebox{\textwidth}{!}{%
\begin{tabular}{@{}c|cccc|cccc@{}}
\toprule
\multirow{2}{*}{Model} & \multicolumn{4}{c}{UMLS} & \multicolumn{4}{|c}{Kinship} \\
 & MRR & MR & Hit@1 & Hit@3 & MRR & MR & Hit@1 & Hit@3 \\ \midrule
PathCon & 0.9385 \footnotesize{$\pm$ 0.0151} & \underline {1.1791 \footnotesize{$\pm$ 0.0531}} & 0.8941 \footnotesize{$\pm$ 0.0246} & 0.9816 \footnotesize{$\pm$ 0.0110} & 0.8931 \footnotesize{$\pm$ 0.0146} & 1.3031 \footnotesize{$\pm$ 0.0298} & 0.8062 \footnotesize{$\pm$ 0.0282} & 0.9807 \footnotesize{$\pm$ 0.0021} \\ \midrule
\ourmodel{}~(L) & 0.9289 \footnotesize{$\pm$ 0.0355} & 1.2475 \footnotesize{$\pm$ 0.1040} & 0.8794 \footnotesize{$\pm$ 0.0584} & 0.9759 \footnotesize{$\pm$ 0.0201} & 0.9004 \footnotesize{$\pm$ 0.0060} & 1.2922 \footnotesize{$\pm$ 0.0185} & 0.8211 \footnotesize{$\pm$ 0.0119} & 0.9791 \footnotesize{$\pm$ 0.0015} \\
\ourmodel{}~(G) & \underline{ 0.9465 \footnotesize{$\pm$ 0.0144}} & 1.1972 \footnotesize{$\pm$ 0.1024} & \underline{ 0.9100 \footnotesize{$\pm$ 0.0204}} & \underline{ 0.9853 \footnotesize{$\pm$ 0.0102}} & 0.9013 \footnotesize{$\pm$ 0.0069} & 1.2894 \footnotesize{$\pm$ 0.0153} & 0.8232 \footnotesize{$\pm$ 0.0137} & 0.9777 \footnotesize{$\pm$ 0.0024} \\
\ourmodel{}~(R) & \textbf{0.9492 \footnotesize{$\pm$ 0.0089}} & \textbf{1.1425 \footnotesize{$\pm$ 0.0375}} & \textbf{0.9106 \footnotesize{$\pm$ 0.0143}} & \textbf{0.9887 \footnotesize{$\pm$ 0.0068}} & 0.8879 \footnotesize{$\pm$ 0.0154} & 1.3233 \footnotesize{$\pm$ 0.0322} & 0.7965 \footnotesize{$\pm$ 0.0301} & 0.9807 \footnotesize{$\pm$ 0.0019} \\
\ourmodel{}~(L+G) & 0.9183 \footnotesize{$\pm$ 0.0208} & 1.2697 \footnotesize{$\pm$ 0.0850} & 0.8622 \footnotesize{$\pm$ 0.0342} & 0.9703 \footnotesize{$\pm$ 0.0206} & 0.9230 \footnotesize{$\pm$ 0.0181} & 1.2061 \footnotesize{$\pm$ 0.0430} & 0.8584 \footnotesize{$\pm$ 0.0356} & \textbf{0.9879 \footnotesize{$\pm$ 0.0039}} \\
\ourmodel{}~(L+R) & 0.8997 \footnotesize{$\pm$ 0.0176} & 1.3572 \footnotesize{$\pm$ 0.0873} & 0.8322 \footnotesize{$\pm$ 0.0293} & 0.9647 \footnotesize{$\pm$ 0.0098} & \multicolumn{1}{l}{\textbf{0.9426 \footnotesize{$\pm$ 0.0065}}} & \multicolumn{1}{l}{\textbf{1.1725 \footnotesize{$\pm$ 0.0144}}} & \multicolumn{1}{l}{\textbf{0.8973 \footnotesize{$\pm$ 0.0144}}} & \multicolumn{1}{l}{{ 0.9869 \footnotesize{$\pm$ 0.0038}}} \\
\ourmodel{}~(G+R) & 0.9369 \footnotesize{$\pm$ 0.0134} & { 1.1819 \footnotesize{$\pm$ 0.0421}} & 0.8928 \footnotesize{$\pm$ 0.0212} & 0.9825 \footnotesize{$\pm$ 0.0075} & 0.8939 \footnotesize{$\pm$ 0.0111} & 1.3086 \footnotesize{$\pm$ 0.0279} & 0.8102 \footnotesize{$\pm$ 0.0210} & 0.9781 \footnotesize{$\pm$ 0.0013} \\
\ourmodel{}~(L+G+R) & 0.9194 \footnotesize{$\pm$ 0.0185} & 1.2106 \footnotesize{$\pm$ 0.0530} & 0.8578 \footnotesize{$\pm$ 0.0314} & 0.9844 \footnotesize{$\pm$ 0.0050} & \underline{ 0.9370 \footnotesize{$\pm$ 0.0118}} & \underline{ 1.1756 \footnotesize{$\pm$ 0.0240}} & \underline{ 0.8859 \footnotesize{$\pm$ 0.0233}} & \underline{0.9873 \footnotesize{$\pm$ 0.0025}} \\ \bottomrule
\end{tabular}
}
\end{table*}

\subsection{Results\label{sec:results42}}
The empirical results are reported in Table~\ref{tab:results1} and Table~\ref{tab:results2} on the large datasets (WN18, WN18RR, FB15k, and FB15k-237), and in Table~\ref{tab:results3} on the small datasets (UMLS, and Kinship). $L$, $G$, and $R$ stand for \ourmodel{} relying on local, global, and random attention mechanisms only, respectively. When there is more than one mechanism taking place, we add two or more of those letters. \ourmodel{} achieves competitive results, outperforming the other methods on all metrics for WN18, WN18RR, UMLS, and Kinship datasets and two out four metrics for FB15k. For FB15k-237, PathCon achieves the higher values on three metrics, yet \ourmodel{} surpassed the other models. These findings go towards our initial premise that contextual information does enhance the learned representations.


In general, when \ourmodel{} uses the global attention only, it achieves the best overall results reaching the best or the second-best cases in seven cases for the large datasets. Following, we see that the random attention mechanism is in six times in the first two positions (either it is the best or the second-best result), followed by the combination of local and global attention mechanisms and the random and global attention mechanisms (they are both fives times in the first two places). While the global mechanism may have pushed forward those results, we can see that there are several cases where the global attention is neither the best nor the second-best choice. At the same time, some other method still wins -- for example, Hit@3 of WN18 and MRR of FB15k-237 of the random mechanism is better than global alone or some of its combinations.

The greater absolute gains on the large datasets occur in WN18 and WN18RR, which are sparse KGs, \ie{}, KGs of low average entities' degree. Those results point out that the attention mechanisms are able to reinforce connective patterns between entities, leading to better representations. The learning entities context assisted by the random attention mechanisms demonstrates better results over denser (higher average entities degree) KGs (FB15k, FB15k-237).

PathCon presents itself as a good choice for FB15k-237. Regarding all the datasets and metrics, it reaches the best results in five of the cases and the second-best results in three cases. However, \ourmodel{} ties with PathCon in two of the five winner situations (MRR of FB15k-237 and Hit@3 of FB15k) and is close in the other two (Hit@1 and Hit@3 of FB15k-237). These results further indicate that adding different patterns of attention favors finding missing relations in KGs.

On the smaller datasets (UMLS and Kinship), the entities' context sparse representation provided by the use of the random attention mechanism in \ourmodel{} shows overall better performance by achieving alone the best results in all four metrics on UMLS and in combination with local attention mechanism in three metrics on Kinship. On the other hand, both datasets differ on the density of relationships, \textit{i.e.}, Kinship has a higher nodes' connection degree than UMLS (see~\ref{sec:appA}). This observation led to the second-best results being related to the global attention mechanism on UMLS and the local attention mechanism on Kinship, once a higher nodes' connection degree eases \ourmodel{} to capture semantics from the local neighborhood. At the same time, learned representations from sparse KGs (\textit{e.g.}, UMLS) benefit from capturing semantics based on the global attention mechanism and the representation of the semantic path. The obtained results show that, on smaller and denser KGs, the sparse representation of the entities' neighborhood reinforces once-neglected information during the message-passing iterations.

\subsection{Ablation Studies\label{sec:ablation}}

\begin{figure}[t]
     \centering
     \begin{subfigure}[t]{0.49\textwidth}
        \centering
        \includegraphics[width=\linewidth]{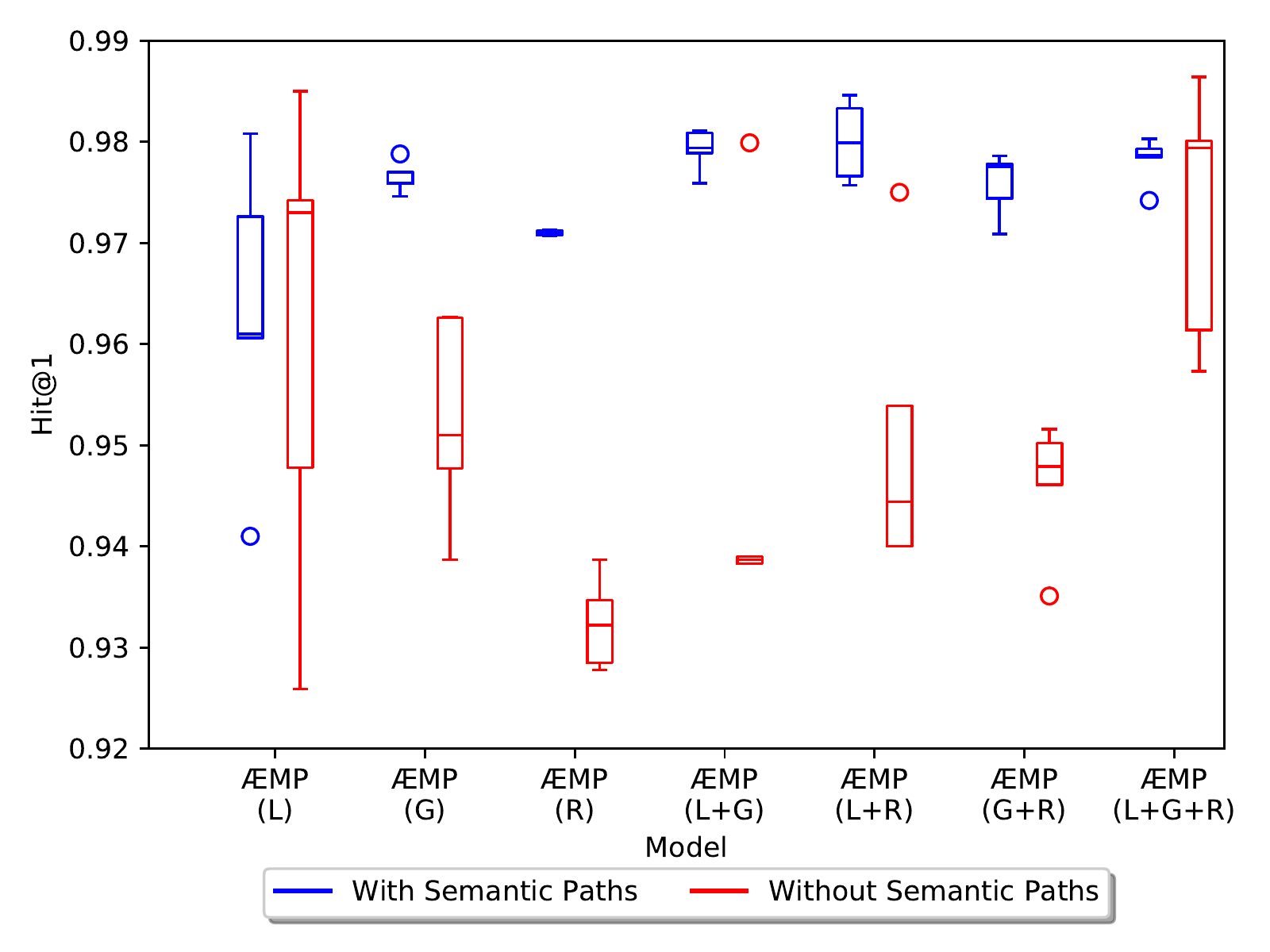}
        \caption{Boxplot of the Hit@1 results from~\ourmodel{} and its subset variations using (or not) the semantic paths' representation to predict new relations.}
        \label{fig:ablation1}
     \end{subfigure}\hfill
     \begin{subfigure}[t]{0.49\textwidth}
        \centering
        \includegraphics[width=\linewidth]{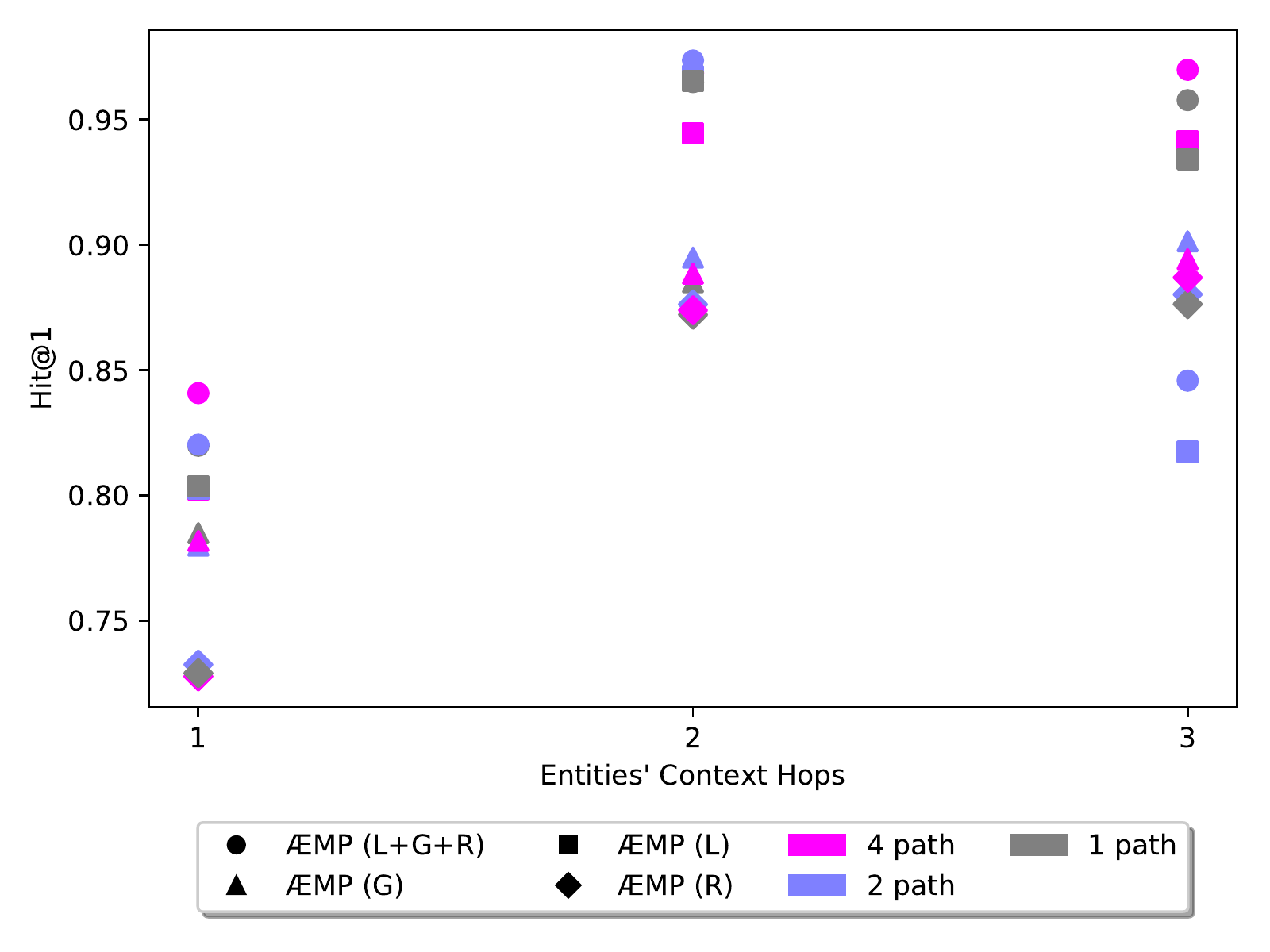}
        \caption{Average Hit@1 results from~\ourmodel{} and its subset variations regarding the variation of entities context hops and semantic path length hyperparameters.}
        \label{fig:ablation2}
     \end{subfigure}
    \caption{Ablation studies \textit{(i)}, and \textit{(ii)}.}
    \label{fig:ablations}
\end{figure}

\begin{figure}[t]
     \centering
     \begin{subfigure}[t]{0.32\textwidth}
        \centering
        \includegraphics[width=\linewidth]{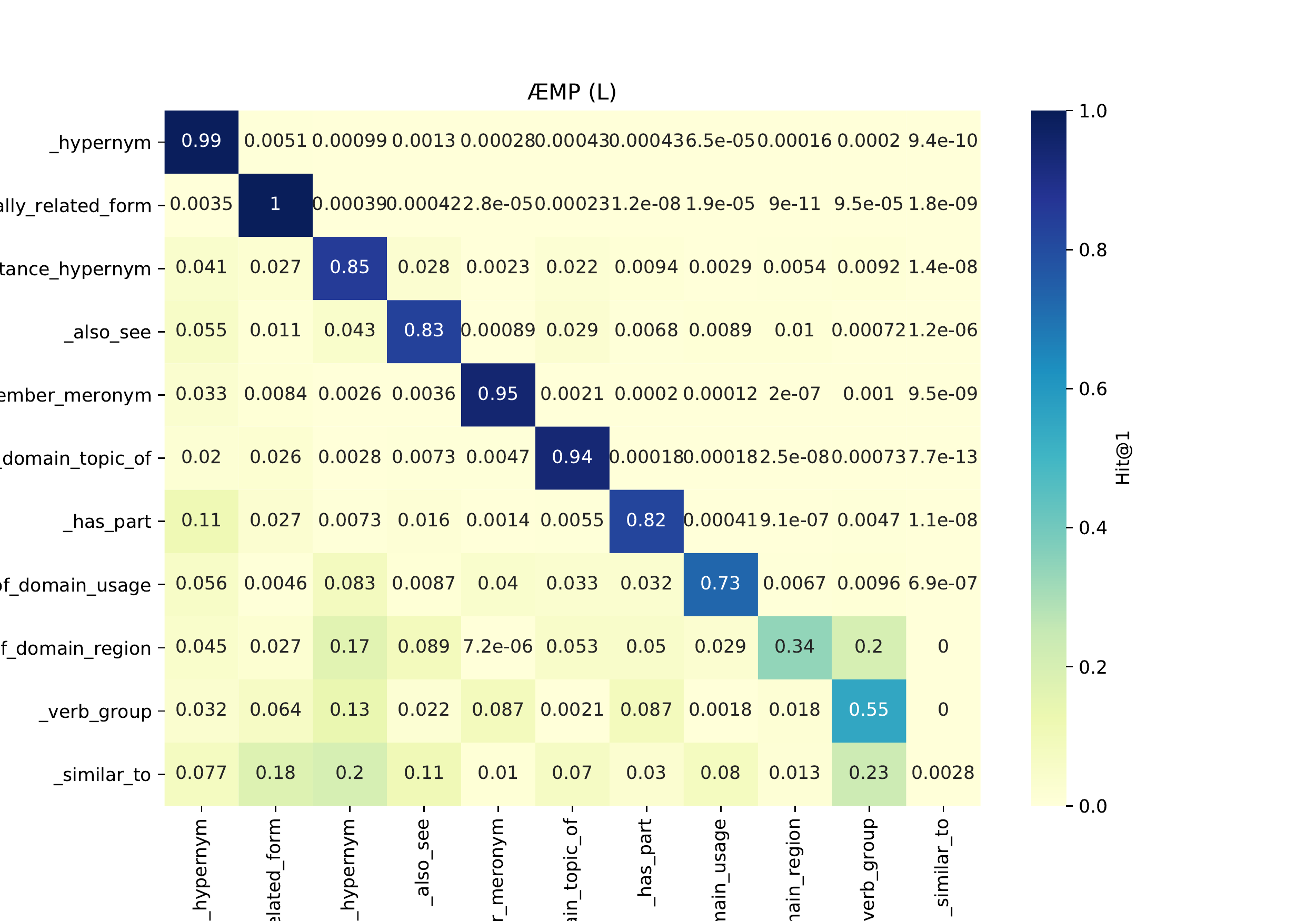}
        \caption{\ourmodel{}~(L)}
        \label{fig:ablation4-latt}
     \end{subfigure}\hfill
     \begin{subfigure}[t]{0.32\textwidth}
        \centering
        \includegraphics[width=\linewidth]{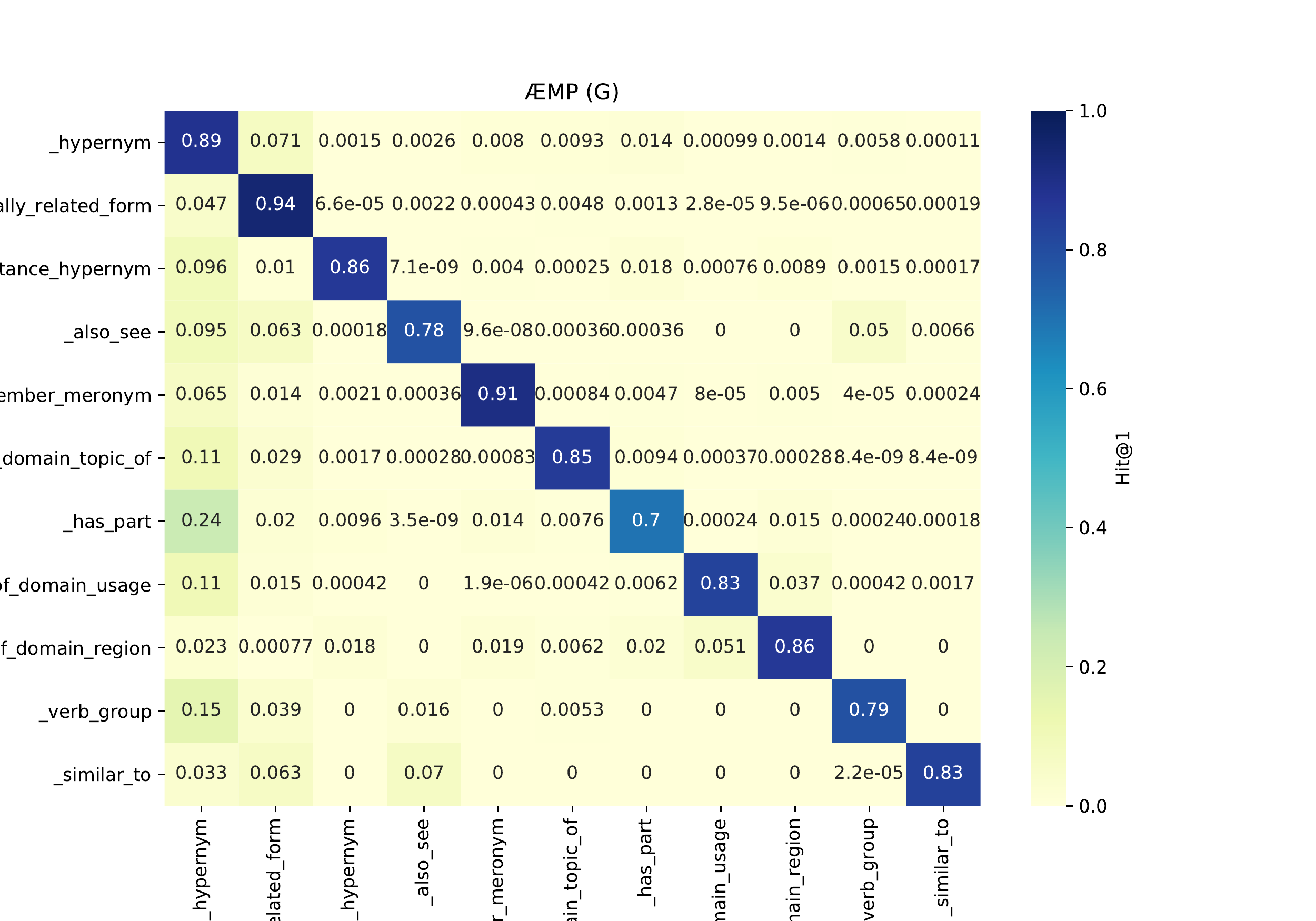}
        \caption{\ourmodel{}~(G)}
        \label{fig:ablation4-gatt}
     \end{subfigure}\hfill
     \begin{subfigure}[t]{0.32\textwidth}
        \centering
        \includegraphics[width=\linewidth]{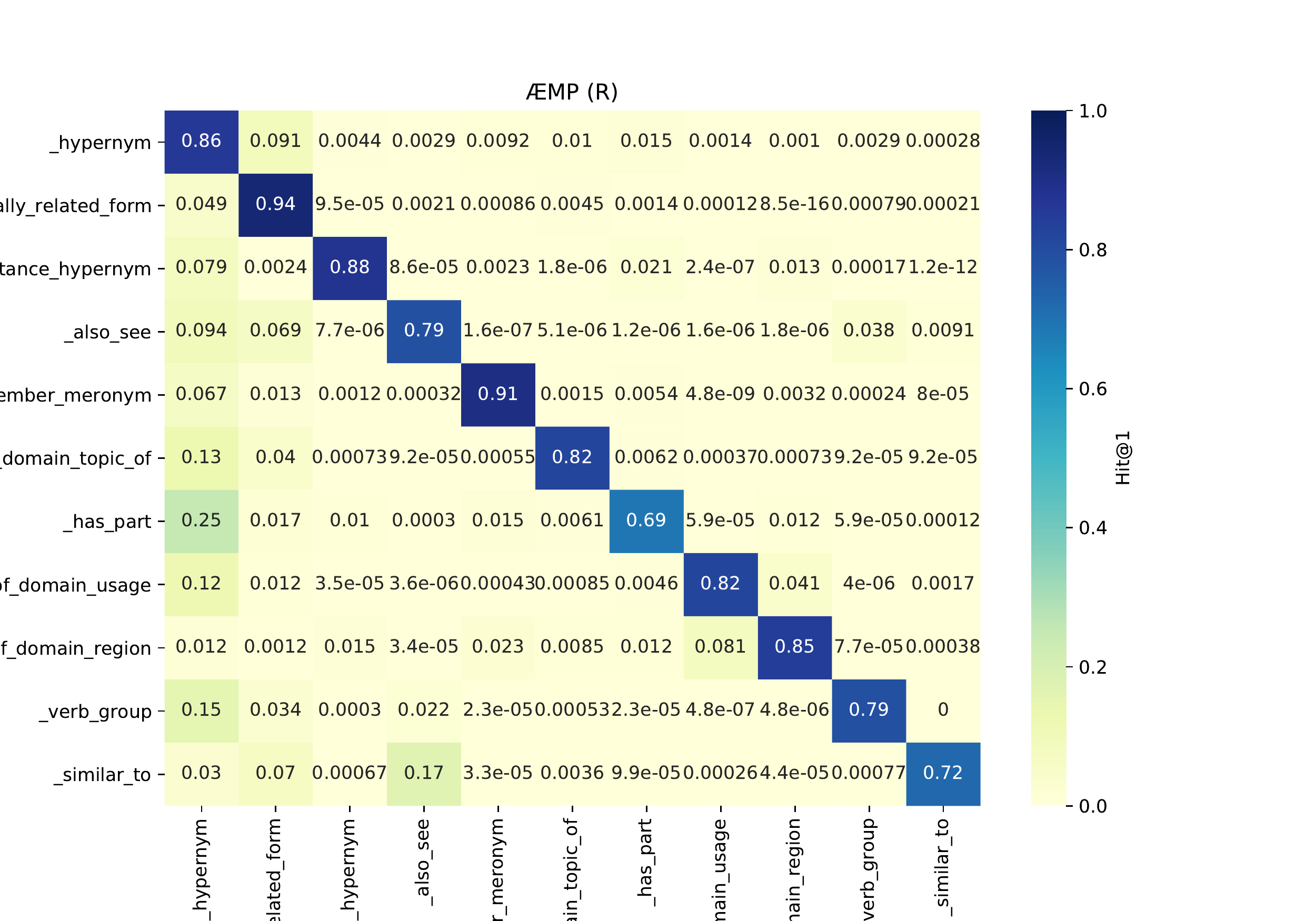}
        \caption{\ourmodel{}~(R)}
        \label{fig:ablation4-ratt}
     \end{subfigure}%
     \vspace{-0.1cm}
    \begin{subfigure}[t]{0.32\textwidth}
        \centering
        \includegraphics[width=\linewidth]{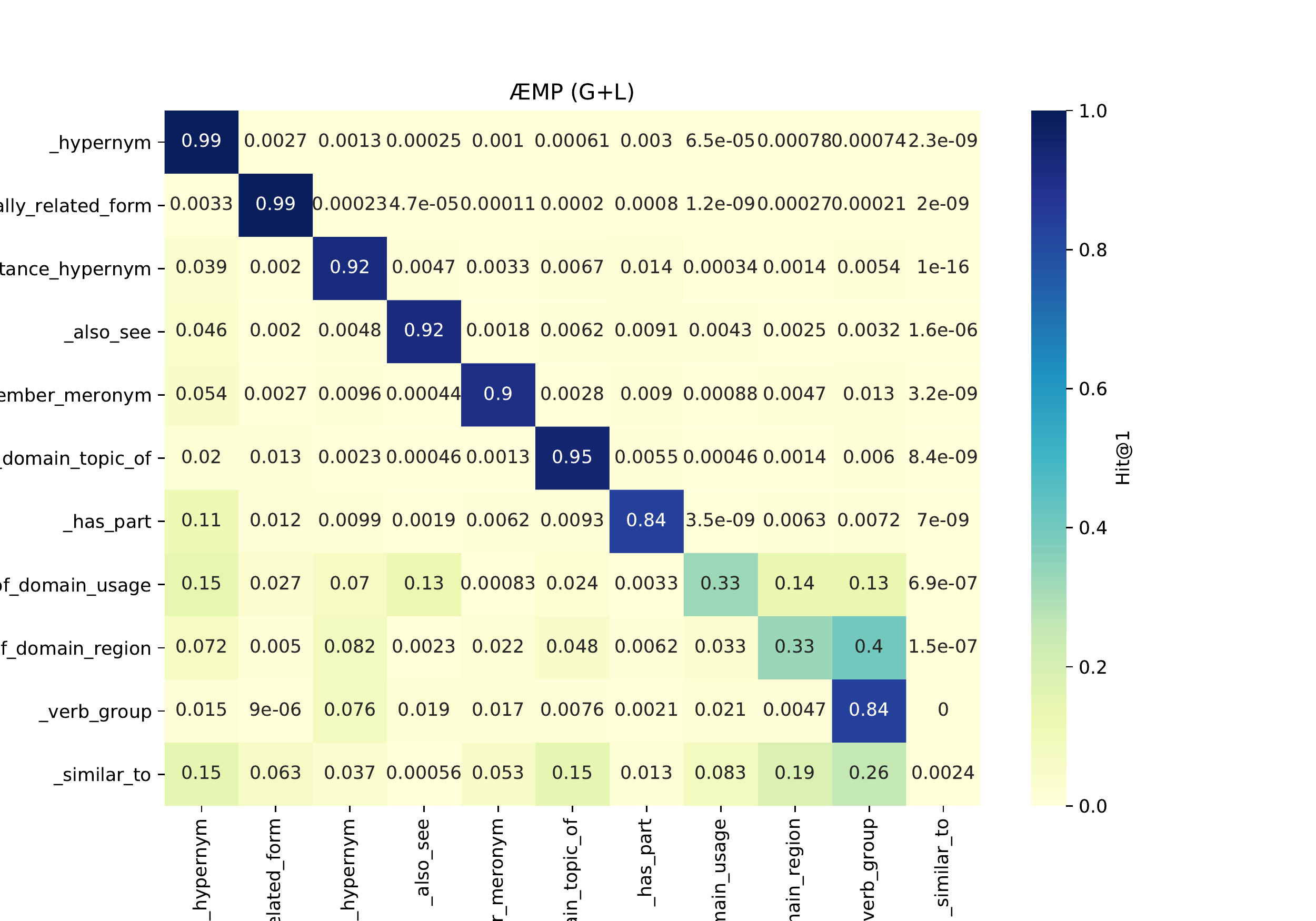}
        \caption{\ourmodel{}~(L+G)}
        \label{fig:ablation4-glatt}
     \end{subfigure}\hfill
     \begin{subfigure}[t]{0.32\textwidth}
        \centering
        \includegraphics[width=\linewidth]{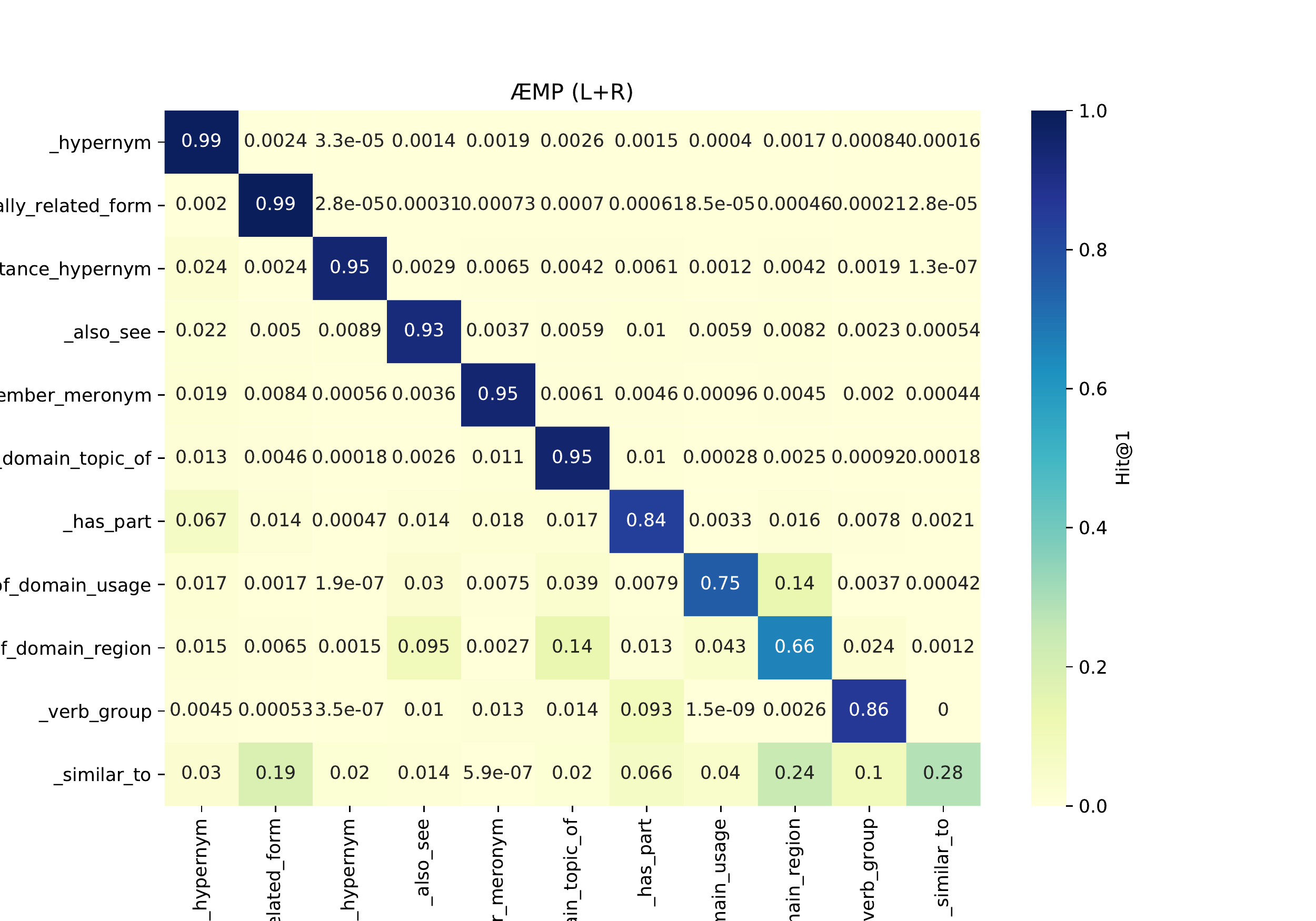}
        \caption{\ourmodel{}~(L+R)}
        \label{fig:ablation4-lratt}
     \end{subfigure}\hfill
     \begin{subfigure}[t]{0.32\textwidth}
        \centering
        \includegraphics[width=\linewidth]{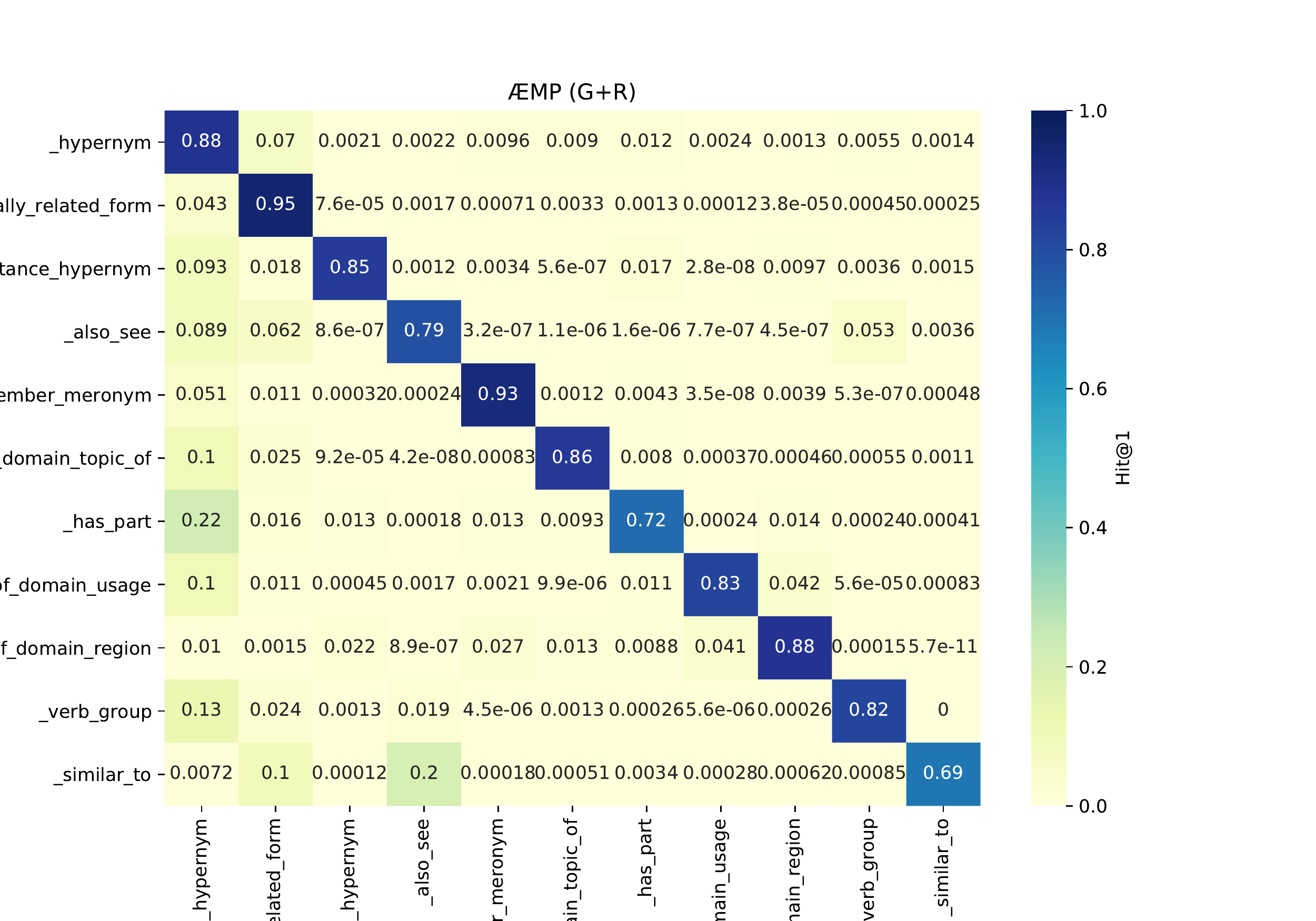}
        \caption{\ourmodel{}~(G+R)}
        \label{fig:ablation4-gratt}
     \end{subfigure}%
     \vspace{-0.1cm}
     \begin{subfigure}[t]{0.32\textwidth}
        \centering
        \includegraphics[width=\linewidth]{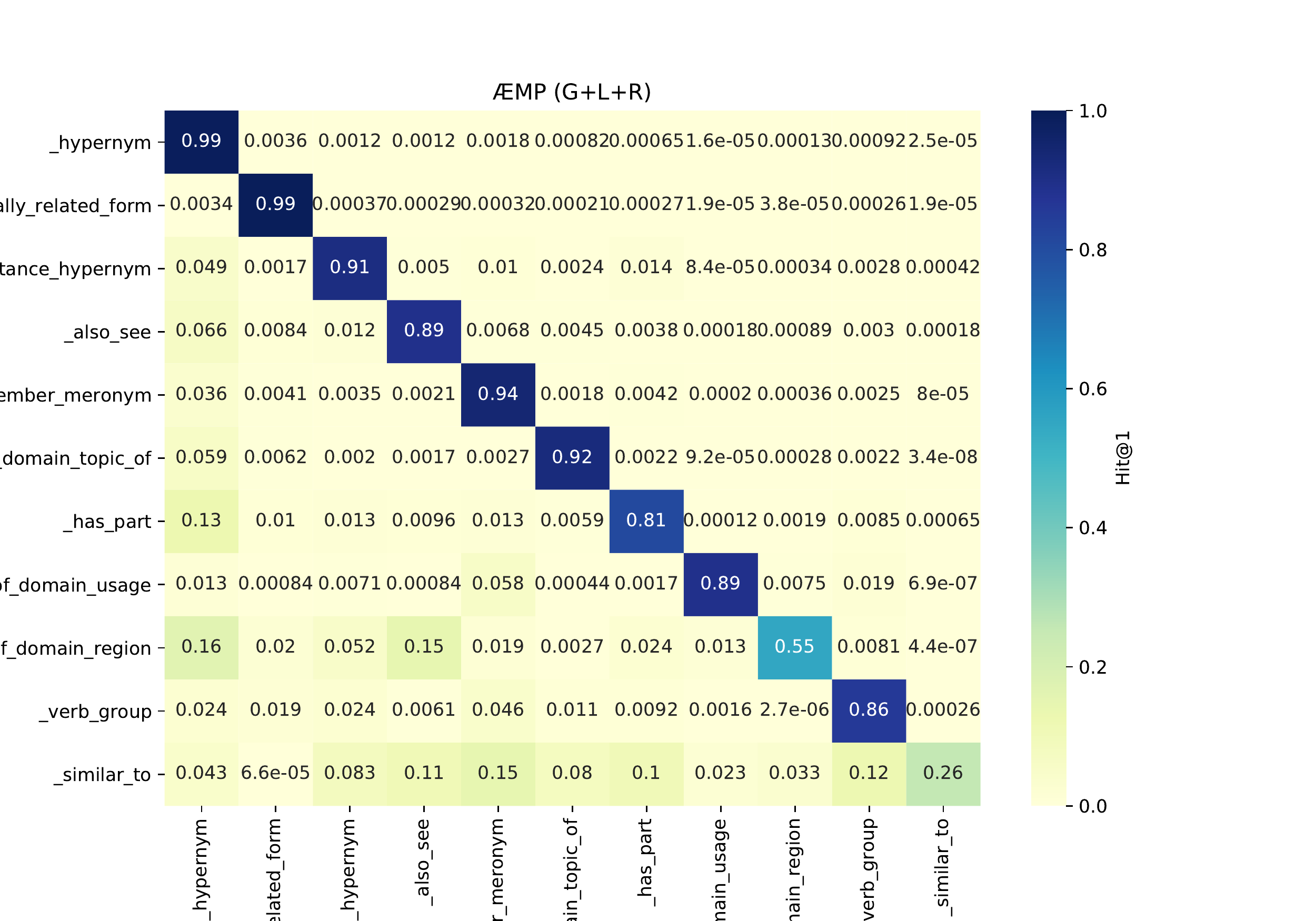}
        \caption{\ourmodel{}~(L+G+R)}
        \label{fig:ablation4-glratt}
     \end{subfigure}
    \caption{Ablation study \textit{(iv)} containing the confusion matrices of \ourmodel{} and its variations on WN18RR. X-axis shows the ground truth and y-axis contains the predicted results. Besides, the axes are ordered in descending order of the relations' frequency (\ie{}, top-bottom for y-axis, and left-right for the x-axis are ordered from the most common to the least common relations). The heatmap indicates the Hit@1 metric varying from $0$ to $1$. $L$, $G$, and $R$ stand for the local, global, and random attention mechanisms, respectively.}
    \label{fig:ablation4}
\end{figure}

To better assess the capabilities of~\ourmodel{} we conduct three ablation studies. As such, we \emph{(i)} evaluate the benefits of using semantic paths' representations towards the capacity of \ourmodel{} to predict relations, and \emph{(ii)} investigate the influence of the number of hops and the number of context neighbors, towards the predictions' results. The results of the first ablation studies are illustrated in Figure~\ref{fig:ablations}, and details of used parametrization are shown in~\ref{sec:appC}. Finally, the third \emph{(iii)} ablation study, illustrated in Figure~\ref{fig:ablation4}, inspects closely the relation between correct predictions made by \ourmodel{} and its variations and the relations' frequency within the KG. The parameters' settings in the fourth study are the same as those used in the previous section.

In the first study (Figure~\ref{fig:ablation1}), we assess the top 1 hit ratio (Hit@1) on WN18RR, varying each attention mechanism and their combination alongside either enabling or disabling the representation of semantic paths. Considering the variations that implement only a subset of the attention mechanisms, \ourmodel{} achieves its best performance when it is able to combine both entities' context representation with the representation of the semantic path. However, when all three local, global, and random attention mechanisms are enabled, and their patterns are observable, \ourmodel{} reaches equivalent results when either using the representation of the semantic paths or not. This observation empirically indicates that the attention-enhanced message-passing scheme might be able to not only represents entities' local neighborhood but also to learn longer sequences of relationships, \ie{}, the semantic paths between the head and tail entities.

In the second study (Figure~\ref{fig:ablation2}), we assess the top 1 hit ratio on WN18RR regarding different numbers of entities' context hops and semantic path length. \ourmodel{} (circle symbol) achieves the overall better results using two contexts hops. In comparison, the subset variations achieve overall better results using the maximum numbers of entities' context hops and semantic path length. Those results reinforce the previous study, suggesting that~\ourmodel{} with its attention-enhanced message-passing scheme have generalizations capabilities, capturing contextual semantics with less context information.

The third study's goal is to investigate the learning capabilities enabled by each attention mechanism and their combinations. To that, we analyze the Hit@1 performance of \ourmodel{} and its variations regarding each existing relation within the WN18RR dataset.
Figure~\ref{fig:ablation4} depicts the confusion matrices of each \ourmodel{} and its variations. The local attention mechanism (\ourmodel{}~(L) -- Figure~\ref{fig:ablation4-latt}) demonstrates better performance predicting commonly seen relations and an almost perfect score on the top two most common relations. However, on rarely seen relations, the method tends to present poor results. In contrast, the global and random attention mechanisms (\ourmodel{}~(G) -- Figure~\ref{fig:ablation4-gatt} and \ourmodel{}~(R) -- Figure~\ref{fig:ablation4-ratt}, respectively) presents smoother results on rarely seen relations, but both perform worst than the \ourmodel{}~(L) on the most common relations. The combination of local attention mechanism with global or local attention mechanisms (\ourmodel{}~(L+G) -- Figure~\ref{fig:ablation4-glatt} and \ourmodel{}~(L+R) -- Figure~\ref{fig:ablation4-lratt}, respectively) presents improvement on rarely seen relations, while keeping the good performance on commonly seen relations. Finally, the three combined attentions mechanisms (\ourmodel{}~(L+G+R) -- Figure~\ref{fig:ablation4-glratt}) suffers on the least seen relation. Nonetheless, the model presents overall better results, achieving good higher hit ratios on most relations.

\section{Conclusion\label{sec:5}}
In this paper, we devised \ourmodel{}, a novel learning method that learns contextualized representations from the combination of distinct views of entities context and semantic path context. 
Our results show that \ourmodel{} has the potential to outperform state-of-the-art models in the relation prediction task. The empirical analysis indicates that the proposed attention-enhanced message-passing scheme can represent not only the entities and their context but also the semantic path context. As future works, we point out three research questions: \emph{(i)} can the proposed message-passing scheme be generalized to represent even longer sequences of relations, replacing the representation of semantic paths?; \emph{(ii)} how \ourmodel{} compares with pure symbolic approaches according to their reasoning skills?; and \emph{(iii)} may other design alternatives (\eg{}, recurrent networks or transformers) better represent longer semantic paths?

\section*{Acknowledgments}
The authors would like to thank the Brazilian Research agencies FAPERJ and CNPq for the financial support. 

\clearpage
\bibliographystyle{elsarticle-num-names} 
\bibliography{cas-refs}

\clearpage
\appendix

\section{Datasets\label{sec:appA}}
Statistics regarding entities and relations of the knowledge graphs described in Section~\ref{sec:expsetting}.

\begin{table}[h]
\centering
\caption{Datasets statistics summary.}
\label{tab:datasets}
\resizebox{\linewidth}{!}{%
\begin{tabular}{@{}c|cccc|ccc@{}}
\toprule
\multirow{2}{*}{Dataset} & \multicolumn{4}{c|}{Edges} & \multicolumn{3}{c}{Entities} \\ \cmidrule(l){2-8} 
 & Training & Validation & Test & Total & Total & Unique & Average degree \\ \midrule
WN18 & 141442 & 5000 & 5000 & 151442 & 40943 & 5 & 7.39 $\pm$ 16.46 \\
WN18RR & 86835 & 3134 & 3034 & 93003 & 40943 & 5754 & 4.54 $\pm$ 8.57 \\
FB15k & 483142 & 59071 & 50000 & 592213 & 14951 & 21 & 79.22 $\pm$ 220.72 \\
FB15k-237 & 272115 & 20466 & 17535 & 310116 & 14541 & 314 & 42.65 $\pm$ 127.70 \\
UMLS & 5216 & 652 & 661 & 6529 & 135 & 0 & 96.72 $\pm$ 87.44 \\
Kinship & 8544 & 1068 & 1074 & 10686 & 104 & 0 & 205.50 $\pm$ 1.67 \\\bottomrule
\end{tabular}%
}
\end{table}

\section{Hyperparameters Search Space\label{sec:appB}}
The hyperparameters search space used in the experiments of Section~\ref{sec:results42}.

\begin{table}[h]
\centering
\caption{Search space of the \ourmodel{}'s hyperparameters.}
\label{tab:hyperparameters}
\begin{tabular}{@{}ll@{}}
\toprule
Hyperparamter & Search space \\ \midrule
\multicolumn{1}{l|}{Batch size} & \{64, 128\} \\
\multicolumn{1}{l|}{Epoch} & \{25, 50\} \\
\multicolumn{1}{l|}{Hidden state dimension} & \{64, 128\} \\
\multicolumn{1}{l|}{L2 regularization weight} & \{$10^{-6}$, $10^{-7}$, $10^{-8}$\} \\
\multicolumn{1}{l|}{Learning rate} & \{$10^{-1}$, $10^{-2}$, $10^{-3}$\} \\
\multicolumn{1}{l|}{Maximum entities context hops} & \{1, 2, 3\} \\
\multicolumn{1}{l|}{Maximum semantic path length} & \{1, 2, 3\} \\
\multicolumn{1}{l|}{Random attention context selection criteria} & \{0.2, 0.25, 0.5, 0.75\} \\ \bottomrule
\end{tabular}
\end{table}

\section{Ablation Studies\label{sec:appC}}
\subsection{Parametrization from Section~\ref{sec:ablation}}

The parametrization used in the ablation studies discussed in Section~\ref{sec:ablation}.

\begin{table}[h]
\centering
\caption{Ablation studies parametrization settings.}
\label{tab:ablationhp}
\begin{tabular}{llll}
\toprule
\multirow{2}{*}{Hyperparamter} & \multicolumn{2}{c}{Ablation Studies} \\ \cline{2-3} 
 & \multicolumn{1}{c|}{\textit{\footnotesize{(i)}}} & \multicolumn{1}{c}{\footnotesize{(ii)}} \\ \hline
\multicolumn{1}{l|}{Batch size} & \multicolumn{1}{l|}{128} & \multicolumn{1}{l}{128} \\
\multicolumn{1}{l|}{Epoch} & \multicolumn{1}{l|}{20} & \multicolumn{1}{l}{10} \\
\multicolumn{1}{l|}{Hidden state dimension} & \multicolumn{1}{l|}{128} & \multicolumn{1}{l}{128}\\
\multicolumn{1}{l|}{L2 regularization weight} & \multicolumn{1}{l|}{$10^{-7}$} & \multicolumn{1}{l}{$10^{-7}$}\\
\multicolumn{1}{l|}{Learning rate} & \multicolumn{1}{l|}{$10^{-3}$} & \multicolumn{1}{l}{$10^{-3}$}\\
\multicolumn{1}{l|}{Maximum entities context hops} & \multicolumn{1}{l|}{3} & \multicolumn{1}{l}{\{1,2,3\}}\\
\multicolumn{1}{l|}{Maximum semantic path length} & \multicolumn{1}{l|}{3} & \multicolumn{1}{l}{\{1,2,4\}}\\
\multicolumn{1}{l|}{Random attention context selection criteria} & \multicolumn{1}{l|}{0.2} & \multicolumn{1}{l}{0.2}\\
\bottomrule
\end{tabular}%
\end{table}

\subsection{Extended ablation studies}

In this section, we conduct two more studies on~\ourmodel{}. The first evaluates the scalability of~\ourmodel{} while the second evaluates the use of a sparse or dense representation for the semantic path representation.

\begin{figure}[h]
    \centering
    \includegraphics[width=0.6\linewidth]{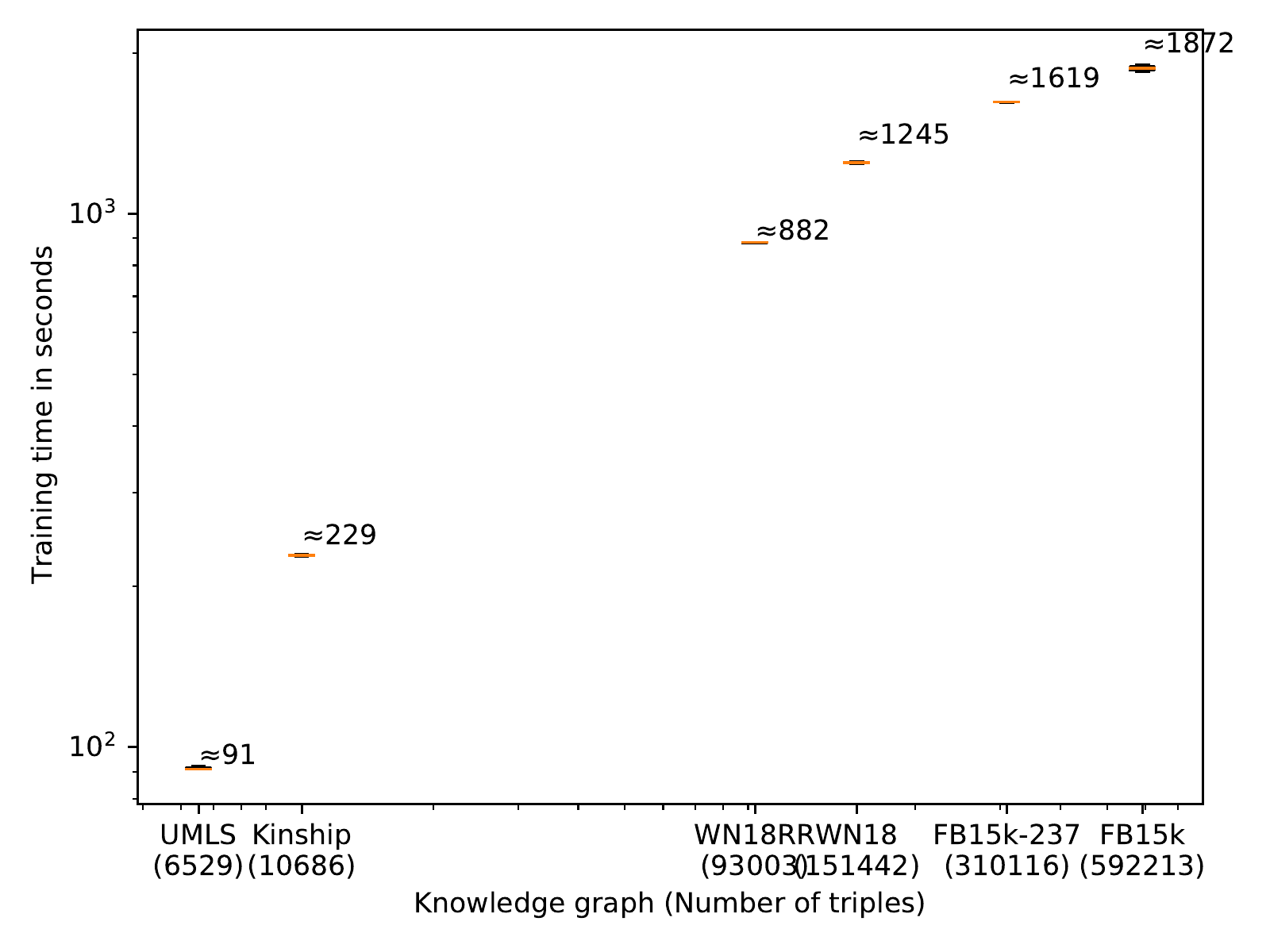}
    \caption{Boxplot of the training time from \ourmodel{} regarding different sized KGs.}
    \label{fig:ablation3}
\end{figure}

The first extended study (Figure~\ref{fig:ablation3}) analyzes the scalability capacity of \ourmodel{}. We measure its training time regarding the number of triples on each previously introduced dataset. Results show that the training time curve of~\ourmodel{} grows slower than linear time over the number of triples, indicating that it is suitable to learn from larger KGs.

\begin{figure}[h]
    \centering
    \includegraphics[width=0.6\linewidth]{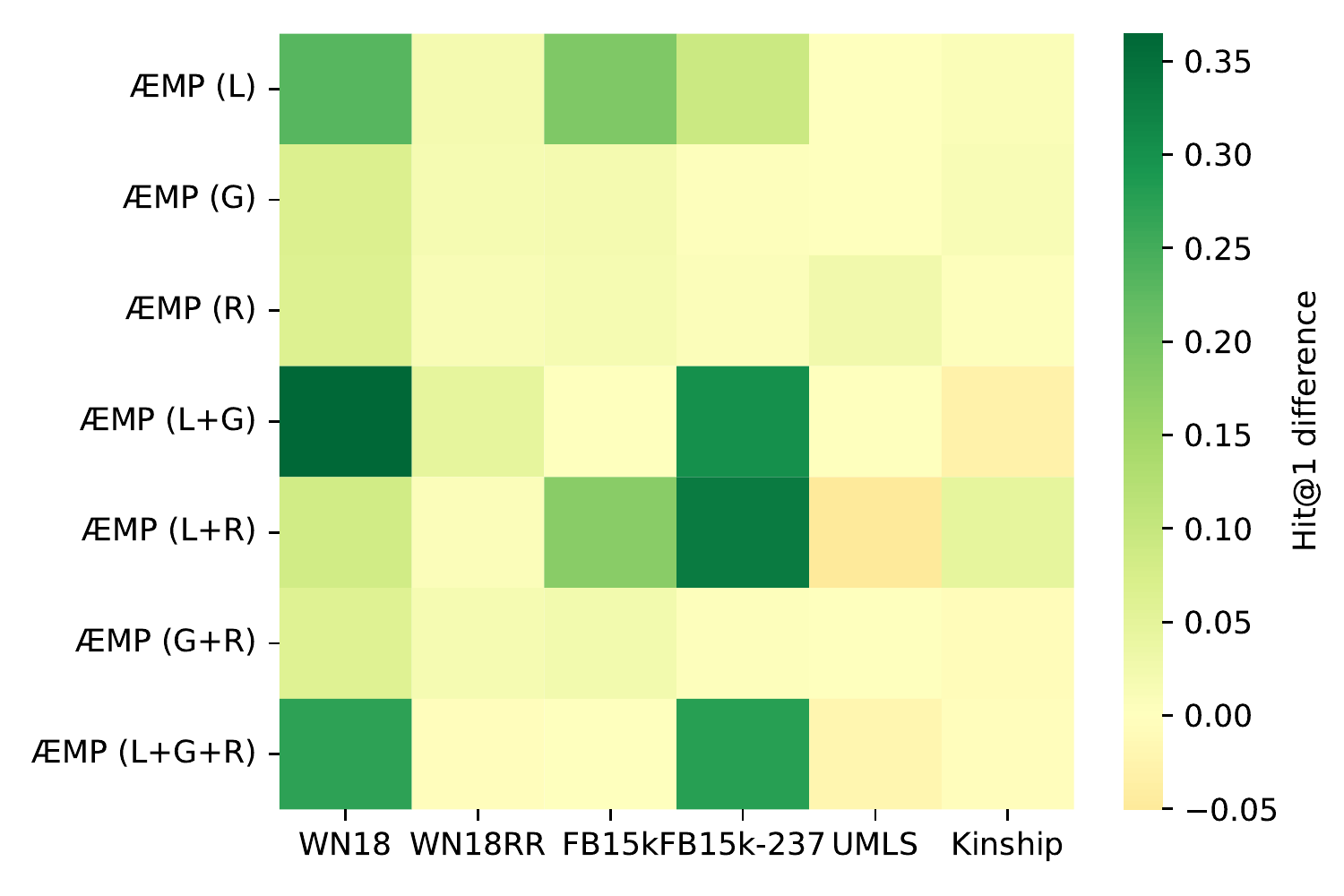}
    \caption{Absolute difference of the best Hit@1 metric between semantic paths sparse and dense representation of each model on each dataset. Zero value means that both representation approaches performs equally, while positive and negative values indicates better performance using sparse and dense representations, respectively.}
    \label{fig:ablation5}
\end{figure}

The second extended study (Table~\ref{tab:rnn3}, and Figure~\ref{fig:ablation5}) discusses the use of a dense representation over a sparse representation to model semantic paths in~\ourmodel{} framework. To do so, we replace the semantic paths capturing process described in~\ref{sec:pathcont} for a recurrent neural network. We measure the MRR and Hit@1 metrics on the six used datasets and insert the results in Table~\ref{tab:rnn3}.

\begin{table*}[h]
\caption{Relation prediction results on all datasets using dense mechanisms to represent semantic paths. $L$, $G$, and $R$ stand for the local, global, and random attention mechanisms, respectively.}
\centering
\resizebox{\textwidth}{!}{%
\begin{tabular}{@{}c|cc|cc|cc@{}}
\toprule
\multirow{2}{*}{Model} & \multicolumn{2}{c|}{WN18} & \multicolumn{2}{c|}{WN18RR} & \multicolumn{2}{c}{FB15k} \\
& MRR & Hit@1 & MRR & Hit@1 & MRR & Hit@1 \\ \midrule
\ourmodel{}~(L)     & 0.7823 \footnotesize{$\pm$ 0.0716} & 0.6689 \footnotesize{$\pm$ 0.1047} & 0.9486 \footnotesize{$\pm$ 0.0281} & 0.9143 \footnotesize{$\pm$ 0.0517} & 0.8226 \footnotesize{$\pm$ 0.0209} & 0.7545 \footnotesize{$\pm$ 0.0228}          \\
\ourmodel{}~(G)     & 0.9554 \footnotesize{$\pm$ 0.0028} & 0.9186 \footnotesize{$\pm$ 0.0051} & 0.9651 \footnotesize{$\pm$ 0.0022} & 0.9379 \footnotesize{$\pm$ 0.0044} & 0.9696 \footnotesize{$\pm$ 0.0003} & 0.9485 \footnotesize{$\pm$ 0.0007}          \\
\ourmodel{}~(R)     & 0.9551 \footnotesize{$\pm$ 0.0033} & 0.9185 \footnotesize{$\pm$ 0.0060} & 0.9626 \footnotesize{$\pm$ 0.0036} & 0.9336 \footnotesize{$\pm$ 0.0065} & 0.9694 \footnotesize{$\pm$ 0.0010} & 0.9480 \footnotesize{$\pm$ 0.0015}          \\
\ourmodel{}~(L+G)   & 0.7616 \footnotesize{$\pm$ 0.1152} & 0.6273 \footnotesize{$\pm$ 0.1466} & 0.9496 \footnotesize{$\pm$ 0.0233} & 0.9182 \footnotesize{$\pm$ 0.0501} & 0.9795 \footnotesize{$\pm$ 0.0303} & 0.9663 \footnotesize{$\pm$ 0.0098} \\
\ourmodel{}~(L+R)   & 0.9243 \footnotesize{$\pm$ 0.0506} & 0.8680 \footnotesize{$\pm$ 0.1024} & 0.9753 \footnotesize{$\pm$ 0.0113} & 0.9603 \footnotesize{$\pm$ 0.0222} & 0.7613 \footnotesize{$\pm$ 0.0143} & 0.6916 \footnotesize{$\pm$ 0.0183}          \\
\ourmodel{}~(G+R)   & 0.9577 \footnotesize{$\pm$ 0.0012} & 0.9234 \footnotesize{$\pm$ 0.0026} & 0.9648 \footnotesize{$\pm$ 0.0016} & 0.9380 \footnotesize{$\pm$ 0.0031} & 0.9688 \footnotesize{$\pm$ 0.0004} & 0.9472 \footnotesize{$\pm$ 0.0007}          \\
\ourmodel{}~(L+G+R) & 0.8417 \footnotesize{$\pm$ 0.0752} & 0.7200 \footnotesize{$\pm$ 0.1562} & 0.9778 \footnotesize{$\pm$ 0.0099} & 0.9655 \footnotesize{$\pm$ 0.0195} & 0.9786 \footnotesize{$\pm$ 0.0017} & 0.9641 \footnotesize{$\pm$ 0.0047} \\ 
\toprule

\multirow{2}{*}{Model} & \multicolumn{2}{c|}{FB15k-237} & \multicolumn{2}{c|}{UMLS} & \multicolumn{2}{c}{Kinship}\vspace{-0.1cm}\\\vspace{-0.1cm}
& \scriptsize{MRR} & \scriptsize{Hit@1} & \scriptsize{MRR} & \scriptsize{Hit@1} & \scriptsize{MRR} & \scriptsize{Hit@1} \\ \midrule
\ourmodel{}~(L)     & 0.7369 \footnotesize{$\pm$ 0.0340} & 0.6268 \footnotesize{$\pm$ 0.0437} & 0.9298 \footnotesize{$\pm$ 0.0180} & 0.8794 \footnotesize{$\pm$ 0.0265} & 0.8954 \footnotesize{$\pm$ 0.0057} & 0.8106 \footnotesize{$\pm$ 0.0114}          \\
\ourmodel{}~(G)     & 0.9752 \footnotesize{$\pm$ 0.0005} & 0.9584 \footnotesize{$\pm$ 0.0009} & 0.9476 \footnotesize{$\pm$ 0.0204} & 0.9091 \footnotesize{$\pm$ 0.0342} & 0.8955 \footnotesize{$\pm$ 0.0086} & 0.8108 \footnotesize{$\pm$ 0.0168}          \\
\ourmodel{}~(R)     & 0.9759 \footnotesize{$\pm$ 0.0007} & 0.9586 \footnotesize{$\pm$ 0.0009} & 0.9299 \footnotesize{$\pm$ 0.0027} & 0.8825 \footnotesize{$\pm$ 0.0072} & 0.8841 \footnotesize{$\pm$ 0.0273} & 0.7906 \footnotesize{$\pm$ 0.0522}          \\
\ourmodel{}~(L+G)   & 0.7600 \footnotesize{$\pm$ 0.0311} & 0.6612 \footnotesize{$\pm$ 0.0366} & 0.9219 \footnotesize{$\pm$ 0.0160} & 0.8603 \footnotesize{$\pm$ 0.0282} & 0.9380 \footnotesize{$\pm$ 0.0066} & 0.8863 \footnotesize{$\pm$ 0.0133}          \\
\ourmodel{}~(L+R)   & 0.6307 \footnotesize{$\pm$ 0.0229} & 0.5040 \footnotesize{$\pm$ 0.0460} & 0.9324 \footnotesize{$\pm$ 0.0191} & 0.8825 \footnotesize{$\pm$ 0.0303} & 0.9176 \footnotesize{$\pm$ 0.0192} & 0.8496 \footnotesize{$\pm$ 0.0344}          \\
\ourmodel{}~(G+R)   & 0.9757 \footnotesize{$\pm$ 0.0004} & 0.9583 \footnotesize{$\pm$ 0.0007} & 0.9359 \footnotesize{$\pm$ 0.0091} & 0.8922 \footnotesize{$\pm$ 0.0120} & 0.8983 \footnotesize{$\pm$ 0.0093} & 0.8160 \footnotesize{$\pm$ 0.0175}          \\
\ourmodel{}~(L+G+R) & 0.7745 \footnotesize{$\pm$ 0.0283} & 0.6841 \footnotesize{$\pm$ 0.0330} & 0.9281 \footnotesize{$\pm$ 0.0216} & 0.8794 \footnotesize{$\pm$ 0.0376} & 0.9385 \footnotesize{$\pm$ 0.0043} & 0.8908 \footnotesize{$\pm$ 0.0084}          \\ \bottomrule
\end{tabular}
}
\label{tab:rnn3}
\end{table*}

Also, to eases comparison with the representation proposed in~\ourmodel{}, Figure~\ref{fig:ablation5} shows the absolute difference of the best Hit@1 metric between representations of each model on each dataset. In general, both representations presents similar results in all model-dataset pairs. Specifically, the results indicate a slight performance gain using sparse representation for larger dataset (such as WN18 and FB15k), and, on smaller datasets (UMLS and Kinship), the gains tends to be neutral.

\end{document}